\documentclass{ieeeaccess}
\usepackage{cite}
\usepackage{amsmath,amssymb,amsfonts}
\usepackage{algorithmic}
\usepackage{graphicx}
\usepackage{textcomp}
\def\BibTeX{{\rm B\kern-.05em{\sc i\kern-.025em b}\kern-.08em
    T\kern-.1667em\lower.7ex\hbox{E}\kern-.125em}}

\graphicspath{ {images/} }
\usepackage{caption}
\usepackage{subcaption}

\usepackage{array}
\newcolumntype{L}[1]{>{\raggedright\arraybackslash}m{#1}}
\newcolumntype{M}[1]{>{\centering\arraybackslash}m{#1}}
\usepackage{hyperref}

\begin{document}
\doi{10.1109/ACCESS.2017.DOI}


\title{A Comprehensive Survey on Visual Question Answering Datasets and Algorithms}

\author{\uppercase{Raihan Kabir}\authorrefmark{1},
\uppercase{Naznin Haque}\authorrefmark{1}, 
\uppercase{Md Saiful Islam\authorrefmark{1, 2}, and Marium-E-Jannat}\authorrefmark{1, 3}}
\address[1]{Department of Computer Science and Engineering, Shahjalal University of Science and Technology, Sylhet, Bangladesh}
\address[2]{Department of Computing Science, University of Alberta, Edmonton, Canada}
\address[3]{Department of Computer Science, University of British Columbia, Kelowna, Canada}


\begin{abstract}
Visual question answering(VQA) refers to the problem where given an image and a natural language question about the image, a correct natural language answer has to be generated. A VQA model has to demonstrate both the visual understanding of the image and the semantic understanding of the question. It also has to possess the necessary reasoning capability to deduce the answer correctly. Consequently, a capable VQA model would lead to significant advancements in artificial intelligence. Since the inception of this field, a plethora of VQA datasets and models have been published. 
In this article, we meticulously analyze the current state of VQA datasets and models while cleanly dividing them into distinct categories and then summarize the methodologies and characteristics of each category.
We divide VQA datasets into four categories:
\begin{itemize}
\item
Available datasets that contain a rich collection of authentic images
\item
Synthetic datasets that contain only synthetic images produced through artificial means
\item
Diagnostic datasets that are specially designed to test model performance in a particular area, e.g., understanding the scene text
\item
KB (Knowledge-Based) datasets that are designed to measure a model's ability to utilize outside knowledge
\end{itemize}
Concurrently, we explore six main paradigms of VQA models:
Fusion is where we discuss different methods of fusing information between visual and textual modalities.
Attention is the technique of using information from one modality to filter information from another.
External knowledge base where we discuss different models utilizing outside information.
Composition or Reasoning, where we analyze techniques to answer advanced questions that require complex reasoning steps.
Explanation, which is the process of generating visual and/or textual descriptions to verify sound Reasoning.
Graph models which encode and manipulate relationships through nodes in a graph.
We also discuss some miscellaneous topics, such as scene text understanding, counting, and bias reduction.
\end{abstract}

\begin{keywords}
Visual question answering, computer vision, natural language processing, multimodal learning, machine learning, deep learning, reinforcement learning, neural network.
\end{keywords}

\titlepgskip=-15pt

\maketitle

\section{Introduction}
The field of computer vision has seen significant advances in recent times. Since the development of neural networks, many computer vision problems such as captioning, object recognition, object detection, and scene classification have seen tremendous progress. Moreover, the advent of social networking sites, especially image-sharing websites, has led to abundant image data. Many efficient methods have been developed to extract and collect this data, making large-scale image datasets possible. Crowd-sourcing has also made it possible to quickly utilize human workers to perform captioning, annotation, etc. The same is true for natural language processing, another field to which VQA belongs. Significant work has been done in tasks such as language modeling, word and sentence representation, POS(Parts of speech) tagging, sentiment analysis, textual question answering, etc.
\par
The problem of VQA is that given any image and any natural language question about that image, a correct natural language answer has to be produced.

\begin{figure}[ht]%
    \centering
    \subfloat[\centering]{{\includegraphics[width=4cm, height=4cm]{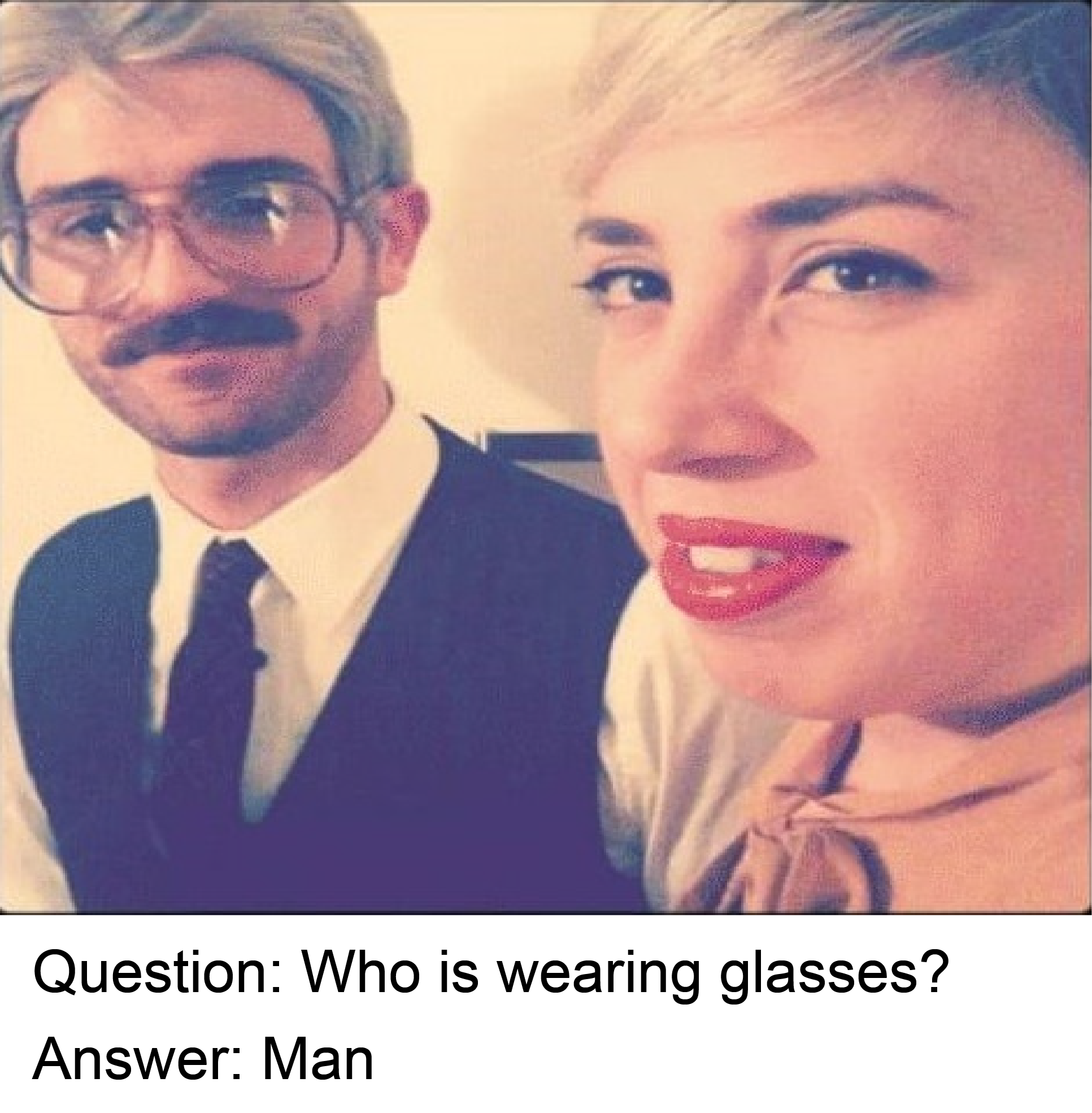} }}%
    \quad
    \subfloat[\centering]{{\includegraphics[width=4cm, height=4cm]{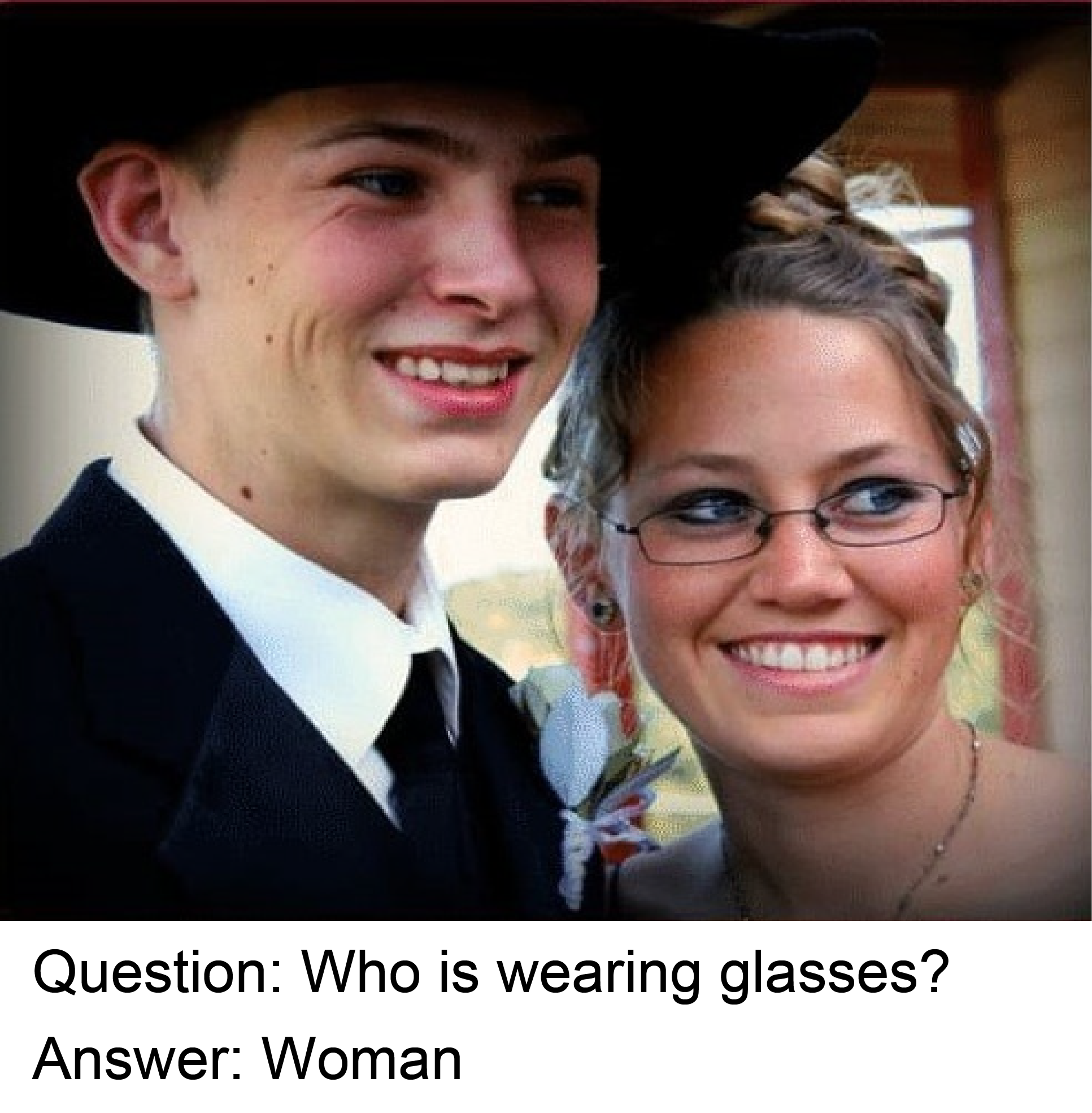} }}%
    \vspace{.5cm}
    \subfloat[\centering]{{\includegraphics[width=4cm, height=4cm]{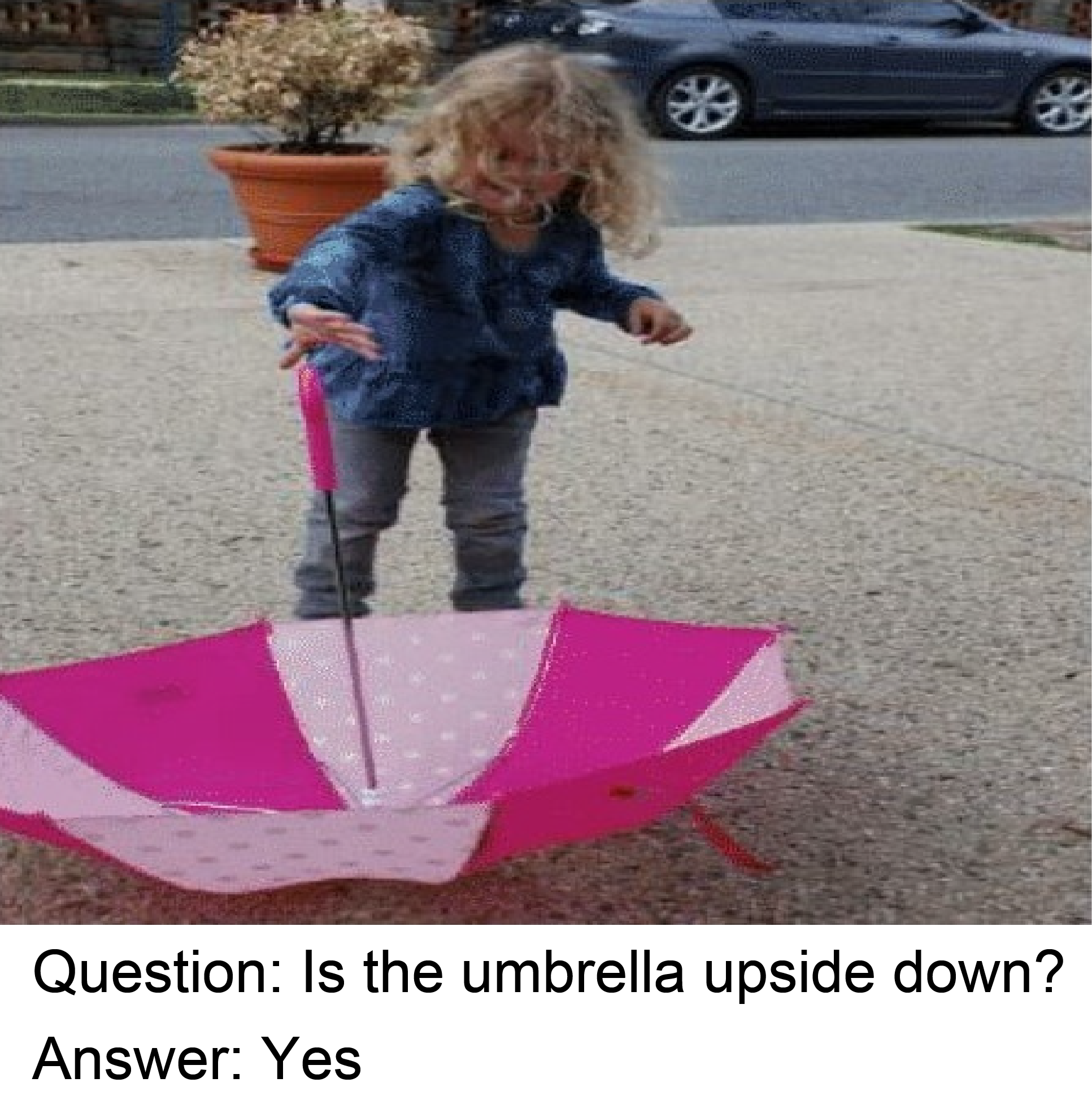} }}%
    \quad
    \subfloat[\centering]{{\includegraphics[width=4cm, height=4cm]{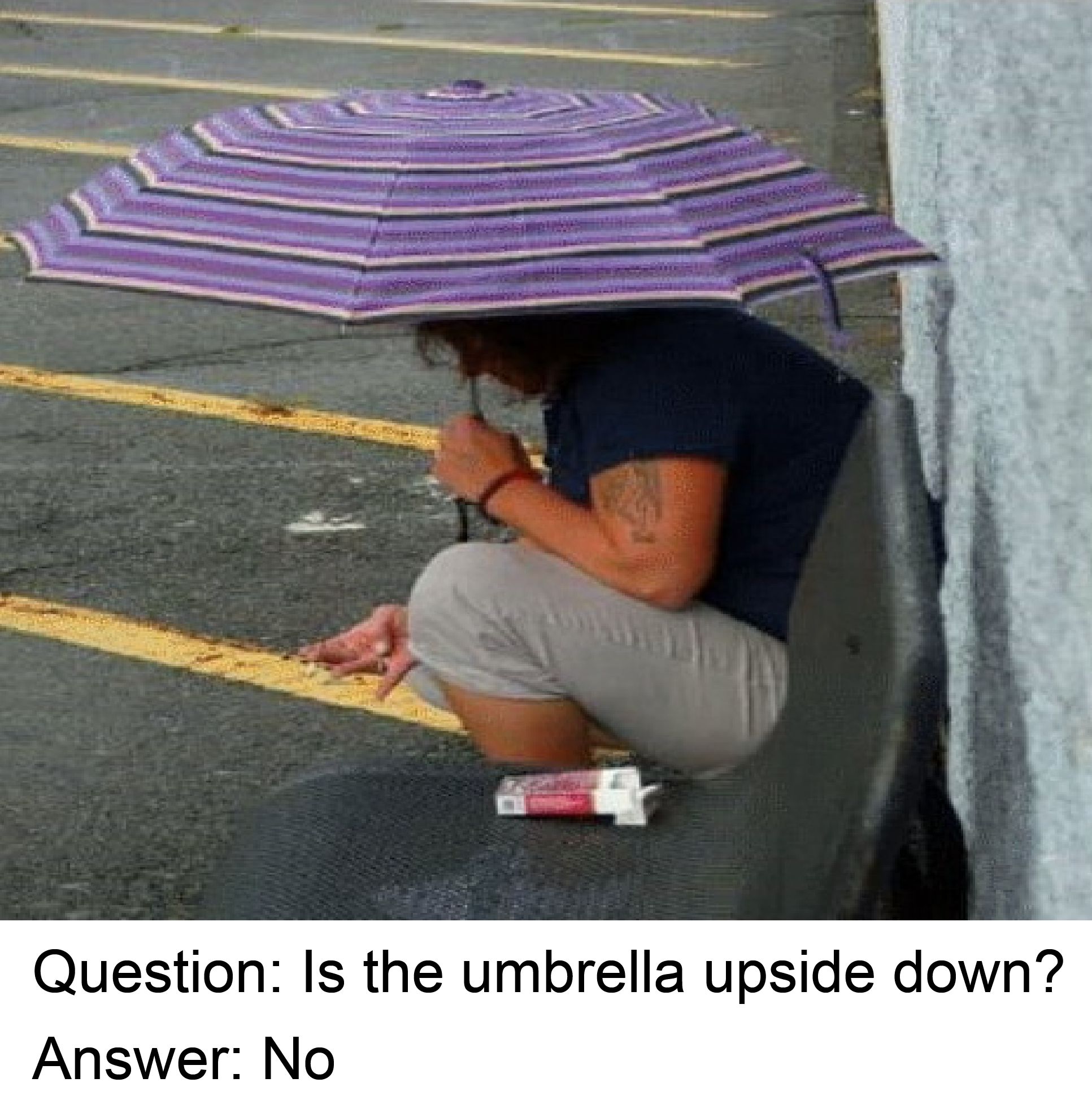} }}%
    \caption{Examples of VQA from the VQA-v2 dataset. Figure from \cite{vqa-v2}}%
\end{figure}

Despite this simple description, VQA encompasses a lot of sub-problems such as:
\begin{itemize}
\item
Object recognition: What is behind the chair?
\item
Object detection: Are there any people in the image?
\item
Counting: How many dogs are there?
\item
Scene classification: Is it raining?
\item
Attribute classification: Is the person happy?
\end{itemize}

In VQA, the image has to be understood with the context given by the question. The same image has to be “seen” differently for different questions. So proper understanding of the question is crucial. Questions can be simple or complex. Sometimes the reasoning process only involves a single object, but it can also involve multiple objects, their attributes, and their relations.
\par
So, what exactly is solving VQA going to achieve? There are many applications where VQA can be applied. An obvious one is to help blind people understand their surroundings. Searching for images can be made easier as the user only has to ask a question, and then image results are generated based on the query. But the ultimate aim of the VQA problem is to pass the Visual Turing Test, which is the ultimate milestone for any AI research. A significant AI must be able to understand and utilize visual information correctly. VQA is a must to test this understanding.

\section{Datasets}
We divide VQA datasets into four categories: 1) General; 2) Synthetic; 3) Diagnostic, and 4) Knowledge-Based. General datasets are large datasets consisting of natural images and human-generated QA pairs. Synthetic datasets contain computer-generated images and questions that are constructed from a fixed number of hand-made templates. Diagnostic datasets are smaller datasets that can be used to isolate and evaluate a specific capability of a model. Finally, knowledge-based datasets provide an external knowledge base that can be queried for extra information.

\subsection{General datasets}
General datasets are the largest, richest, and most used datasets in VQA. General datasets contain many thousands of real-world images from mainstream image datasets like MS-COCO\cite{mscoco} and Imagenet\cite{imagenet}. These datasets are notable for their large scope and diversity. This variety is important as VQA datasets need to reflect the general nature of VQA. Although these datasets do not necessarily capture the endless complexity and variety of visuals in real life, they achieve a close approximation.
\subsubsection{COCO-QA}
COCO-QA was one of the first VQA datasets. It consists of 123K images from MS-COCO. QA pairs were automatically generated from image descriptions. The dataset contains very basic questions and only a few question types. Questions are mainly about object presence, number, color, and location.

\subsubsection{VQA-v1}

\begin{figure}[ht]
\centering
\includegraphics[height=5cm, keepaspectratio]{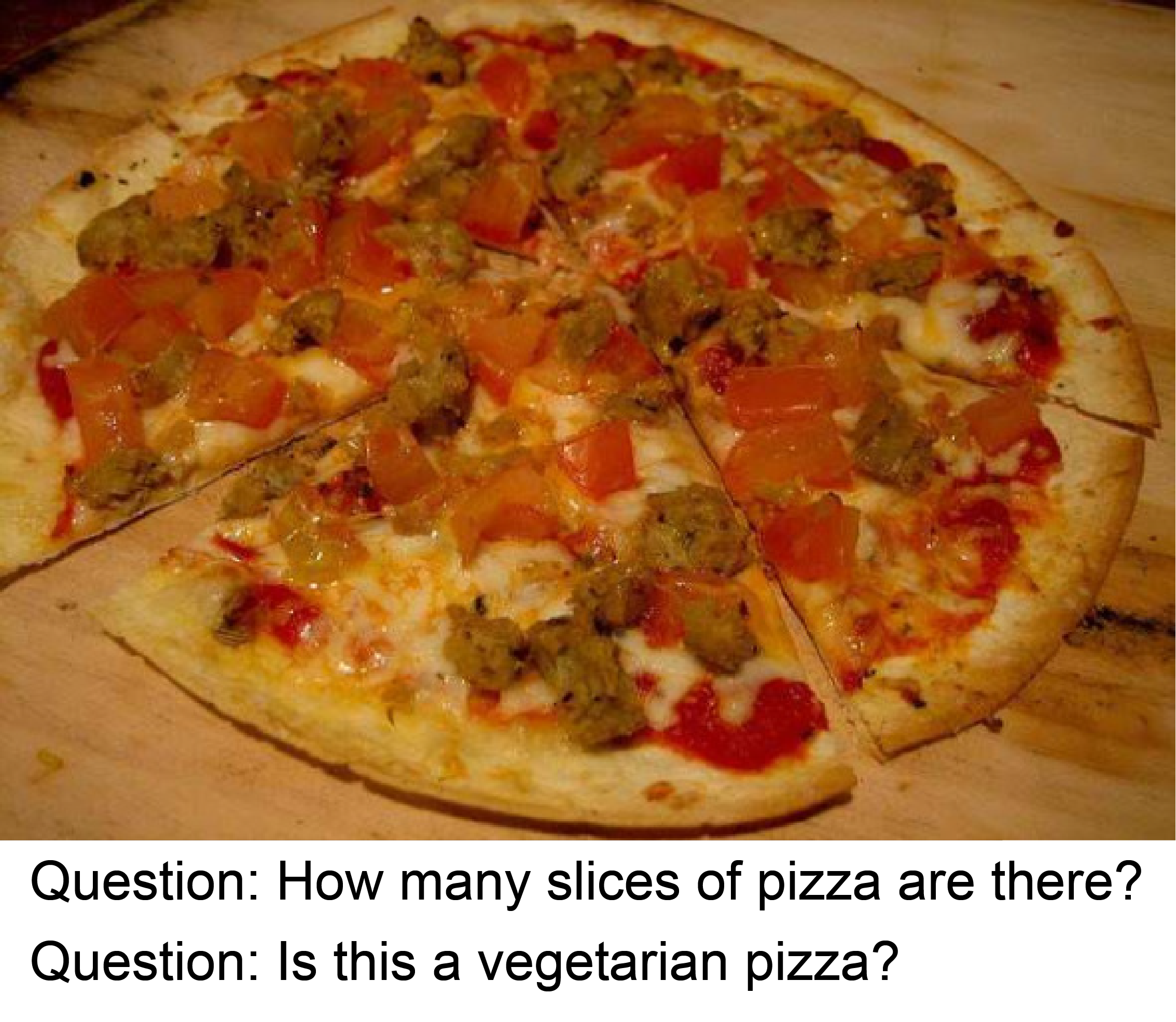}
\caption{An example from VQA-v1\cite{vqa-v1}}
\end{figure}

VQA-v1\cite{vqa-v1}(often called the “VQA dataset”) was the first significantly large VQA dataset. It is divided into two parts, real and abstract. VQA-real consists of real-life images containing multiple objects and a rich, complex mixture of visual information. In contrast, VQA-abstract contains simple artificial scenes generated using clipart software. We talk about vqa-abstract in the synthetic section. Questions and answers were generated using Amazon Mechanical Turk(AMT) workers who were tasked with asking questions that will stump \textit{a smart robot}. There are severe biases in the dataset. For example, a bias of 55.86\% for “yes” in “yes/no” questions and “2” being the most popular answer for “number” questions\cite{vqa-v1}. Many questions also have a low level of inter-human agreement. Many questions ask for subjective answers and opinions which can not be evaluated for correctness. 

\subsubsection{VQA-v2}
VQA-v2\cite{vqa-v2}(also known as balanced VQA) tries to balance the biases in the original VQA-v1 dataset. The authors believed that the original VQA dataset contained significant biases and language priors which enabled models to improve their accuracy by simply leveraging those biases without actually looking at the image. VQA-v2 tries to counter this by having two similar images with two different correct answers for the same question. As a result, the model is forced to look at the image to deduce the correct answer.

\begin{figure}[ht]%
    \centering
    \subfloat[\centering]{{\includegraphics[width=4cm, keepaspectratio]{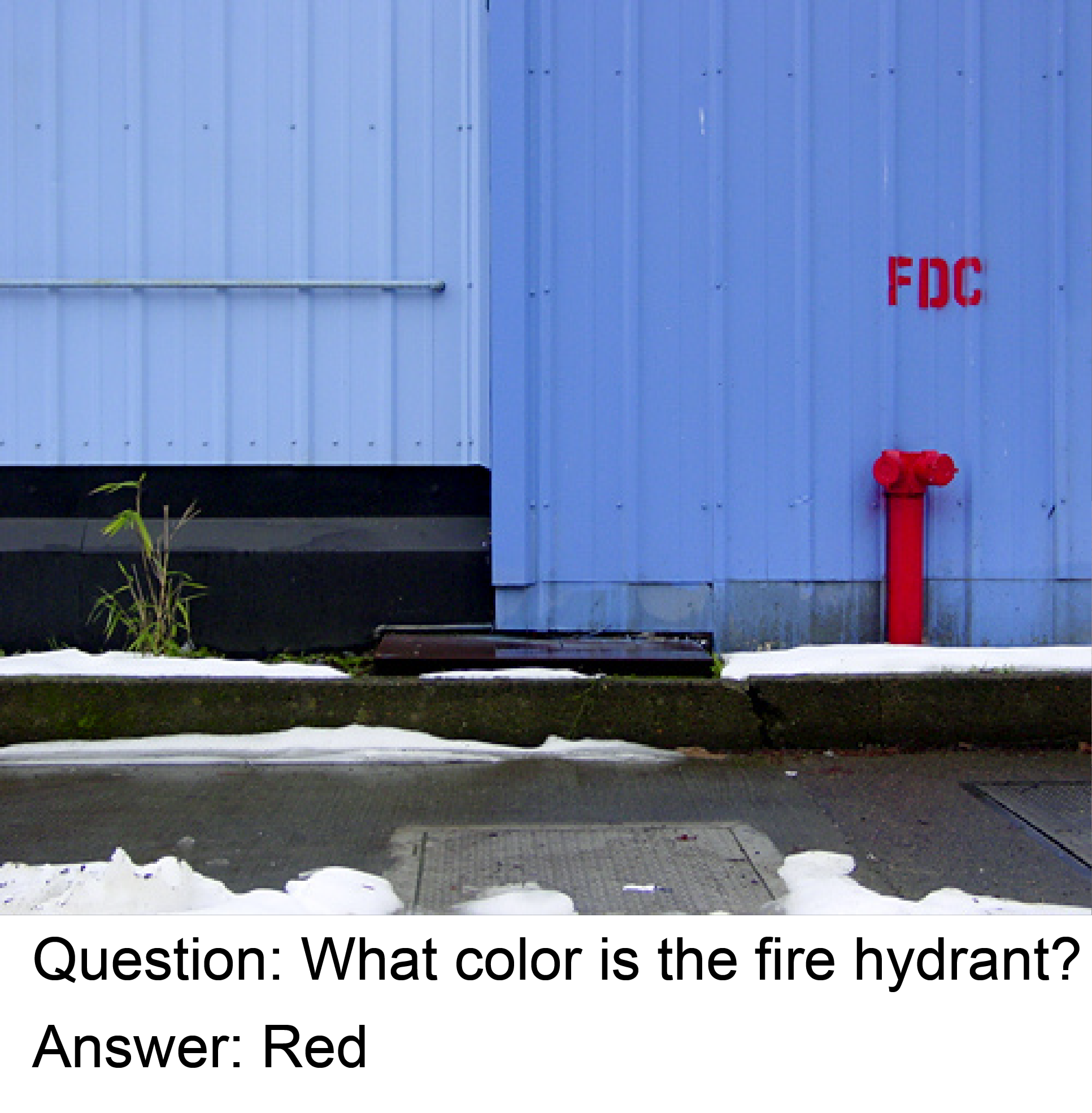} }}%
    \quad
    \subfloat[\centering]{{\includegraphics[width=4cm, keepaspectratio]{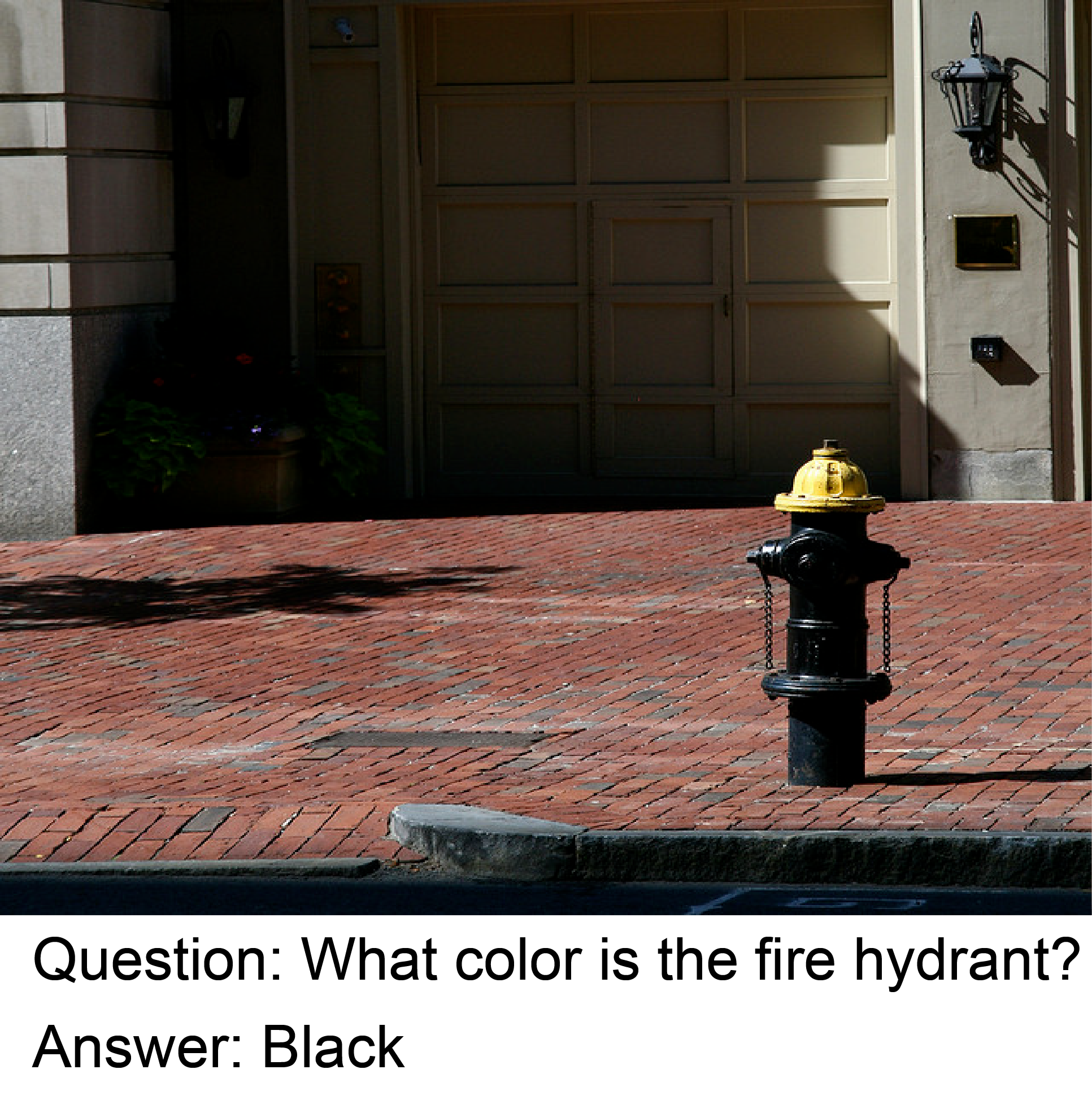} }}%
    \caption{Two complimentary images with different answers for the question “What color is the fire hydrant?” from VQA-v2\cite{vqa-v2}}%
\end{figure}

\subsubsection{Visual genome}
Among the VQA datasets, visual genome(VG)\cite{visualgenome} is one of the richest in terms of information. It contains over 100K images where each image has an average of 21 objects, 18 attributes, and 18 pairwise relationships between objects. The Visual Genome dataset consists of seven main components: 1) region descriptions; 2) objects; 3) attributes; 4) relationships; 5) region graphs; 6) scene graphs; and  7) QA pairs. It is the first VQA dataset to provide a formalized structural representation of an image through scene graphs. Questions were collected specifically so that they were not ambiguous or speculative but answerable if and only if the image is shown. A notable difference between Visual Genome's question distribution and VQA-v1’s is that the authors focus on ensuring that all categories of questions are adequately represented while in VQA-v1, 32.37\% of the questions are “yes/no” binary questions. Visual genome actually excludes binary questions in order to encourage more difficult QA pairs.

\begin{figure}[ht]
     \centering
     \begin{subfigure}[b]{0.2\textwidth}
         \centering
         \includegraphics[width=\textwidth]{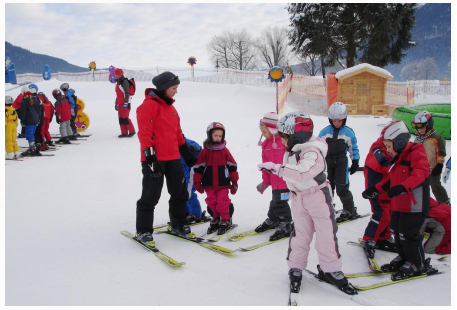}         \caption{}
     \end{subfigure}
     \begin{subfigure}[b]{0.2\textwidth}
         \centering
         \includegraphics[width=1.25\textwidth]{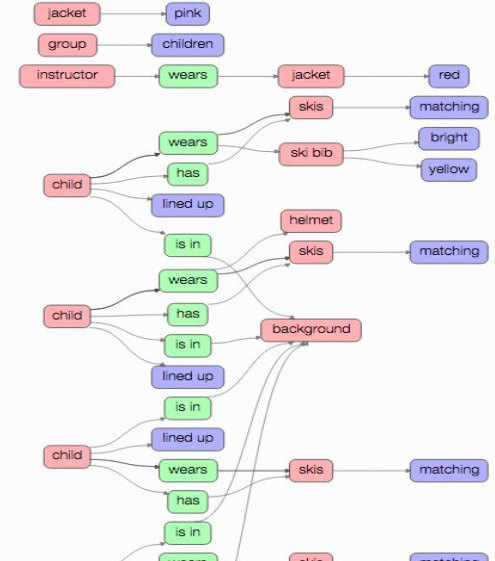}
         \caption{}
     \end{subfigure}
     \hfill
     \caption{An example image(a) from visual genome\cite{visualgenome} with part of it's scene graph(b)}
\end{figure}

\subsubsection{Visual7W}
Visual7W\cite{visual7w} establishes a semantic link between textual descriptions and image regions through object-level grounding. For each word that indicates an object in the question, the bounding boxes of the corresponding object in the image are provided. This helps the model to better associate objects mentioned in the question with objects present in the image. It also enables the model to provide visual answers. Visual7W provides 7 types of questions: what, where, when, who, why, how, and which. The first 6 question types form the “telling” questions which require descriptive answers while the “which” category forms the “pointing” questions that have to point out an object in the image as the answer. “Yes/No” questions are not present in this dataset.

\begin{table*}[t]
\centering
\begin{tabular}{ |M{1cm}|M{1cm}|M{1cm}|M{1cm}|M{2cm}|M{1cm}|M{1cm}|M{2cm}|M{3cm}| } 
 \hline
Dataset & Number of Images & Number of QA Pairs & Question Type & Question Category & Image Source & QA Source & Evaluation Metric & Comments \\
\hline
COCO-QA\cite{coco-qa} & 123K & 118K & OE & Object, Color, Number, and Location & MS COCO & Auto & WUPS & 
Very basic questions, Only a few question types
\\
\hline
 Vqa-v1\cite{vqa-v1} & 204K & 614K(3 per image) & OE and MC & Yes/No, Number, and Other & MS COCO & Manual & vqa metric & 
 Severely biased
 \\
 \hline
 Vqa-v2\cite{vqa-v2} & 200K & 1.1M & OE and MC & Yes/No, Number and Other & MS COCO & Manual & vqa metric &
Slightly less biased
 \\
 \hline
 Visual Genome\cite{visualgenome} & 100K & 1.7M(17 per image) & OE & what, where, how, when, who, and why & MS COCO & Manual & N/A &
Rich annotation
 \\
 \hline
 Visual7W\cite{visual7w} & 47K & 328K & MC & Telling(what, where, how, when, who, why) and pointing(which) & MS COCO & Manual & N/A &
New pointing type question, Object grounding
 \\
 \hline
 TDIUC\cite{tdiuc} & 167K & 1.65M & OE & 12 types & MS COCO & Manual and auto & A- MPT and H-MPT(normalized and unnormalized) &
Better evaluation metrics, Absurd questions
 \\
 \hline
 GQA\cite{gqa} & 113K & 22M & OE & 25 types & MS COCO & Auto & Consistency, validity and plausibility, distribution, grounding &
Better evaluation metrics, Rich structural image and question annotations, Object grounding, Explanation annotation
\\
\hline
\end{tabular}
\caption{Statistics of general VQA datasets.}
\end{table*}

\subsubsection{TDIUC}
Task Driven Image Understanding Challenge(TDIUC)\cite{tdiuc} is specifically tailored to remove biases that plagued previous datasets. It contains about 1.6 million questions about 160K images. The questions are divided into 12 distinct categories each of which represents a specific vision task that a good VQA model should be able to solve. Previous models had severe biases in their distribution of different question types and in the distribution of answers within each question type. TDIUC tries to mitigate these biases as much as possible. It also proposes new evaluation metrics which better evaluate a model's ability to answer each question type. In previous models, doing well on some classes of questions was rewarded more than others. For example, in VQA-v1, improving accuracy on “Is/Are” questions by 15\% will increase overall accuracy by over 5\% but answering all “Why/Where” questions correctly will increase overall accuracy by only 4.1\%. TDIUC also introduces \textit{absurd questions} which should be answered with \textit{does not apply}. It should be noted however that there are only about a thousand unique answers due to most answers being restricted to one or two words.

\subsubsection{GQA}
GQA\cite{gqa} is a dataset that marries compositional reasoning with real-world images. It provides a wealth of data consisting of 22M questions for 113k images. Like Visual Genome, it provides rich, structural, formalized image information through scene graphs. But unlike Visual Genome and like CLEVR\cite{clevr}, it also provides formal, structural representations of questions in the form of functional programs that detail the reasoning steps needed to answer that question. GQA is richer than CLEVR in terms of the variety of object classes and properties as it uses real images instead of synthetic ones. Although GQA questions are automatically generated, they are not completely synthetic like CLEVR. Grammatical rules learned from natural language questions and the use of probabilistic grammar makes GQA questions more human-like and more immune to memorization of language priors. Moreover, sampling and balancing are performed to reduce skews and biases in the question and answer distribution. As questions are not generated by humans, many human-produced biases are not present. Finally, GQA provides new evaluation metrics which better facilitate evaluating a model's true capability.

\begin{table*}[t]
\centering
\begin{tabular}{ |M{2cm}|M{1cm}|M{1cm}|M{1cm}|M{2cm}|M{1cm}|M{2cm}| } 
\hline
Dataset & Number of Images & Number of QA pairs & Type & Composition & QA Source & Evaluation Metric \\
\hline
CLEVR\cite{clevr} & 100K & 1M & OE & Exist, count, compare integer, query attribute and compare attribute & auto & Exact matching \\
\hline
VQA-abstract\cite{vqa-v1} & 50K & 150K & OE and MC & Yes/no, number and other & manual & Vqa metric \\
\hline
SHAPES\cite{nmn} & 15.5K & 15.5K & binary & Yes/no & auto & Exact matching \\
\hline
Yin and yang\cite{yinyang} & 15K & 33K & binay & Yes/no & manual & Vqa metric \\
\hline
\end{tabular}
\caption{Statistics of synthetic VQA datasets.}
\end{table*}

\subsection{Synthetic datasets}
Synthetic datasets contain artificial images, produced using software, instead of real images. A good VQA model should be able to perform well on both real and synthetic data like humans do. Synthetic datasets are easier, less expensive, and less time-consuming to produce as the building of a large dataset can be automated. Synthetic datasets can be tailored so that performing well on them requires better reasoning and composition skills.

\subsubsection{VQA Abstract}
VQA abstract is a subset of the VQA-v1 dataset consisting of 50K artificial scenes. The images were generated using clip-art software. VQA abstract scenes are relatively simple compared to its real counterpart and contain less variety and noise. It was designed this way so a model could focus more on high level reasoning than low level vision tasks.

\subsubsection{Yin and Yang}
Like VQA-v2 tried to balance VQA-v1, Yin and Yang\cite{yinyang} tries to balance the VQA-abstract dataset. They only focus on the yes/no questions and show that models have been using language biases to gain accuracy. As a solution they balance the number of “yes” and “no” questions. They also provide complimentary scenes so that each question has a “yes” answer for one scene and “no” for the other.

\subsubsection{SHAPES}
SHAPES is a small dataset introduced in \cite{nmn}, consisting of 64 images that are composed by arranging colored geometric shapes in different spatial orientations. Each image has the same 244 “yes/no” questions resulting in 15,616 questions. Due to it's small size, it can only be used for diagnostic purposes to evaluate a model's spatial reasoning skills.

\subsubsection{CLEVR}
CLEVR\cite{clevr} is a dataset built solely to evaluate a model's high level reasoning ability. CLEVR contains 100k synthetic images and 1M artificially generated questions. Scenes are presented as collections of objects annotated with shape, size, color, material, and position on the ground-plane. The CLEVR universe contains three object shapes that come in two absolute sizes, two materials, and eight colors. Objects are spatially related via four relationships: “left”, “right”, “behind” and “in front”. CLEVR questions are automatically generated from hand-made templates.
\par

\begin{figure}[ht]
\centering
\includegraphics[width=8cm,
  height=7cm,
  keepaspectratio]{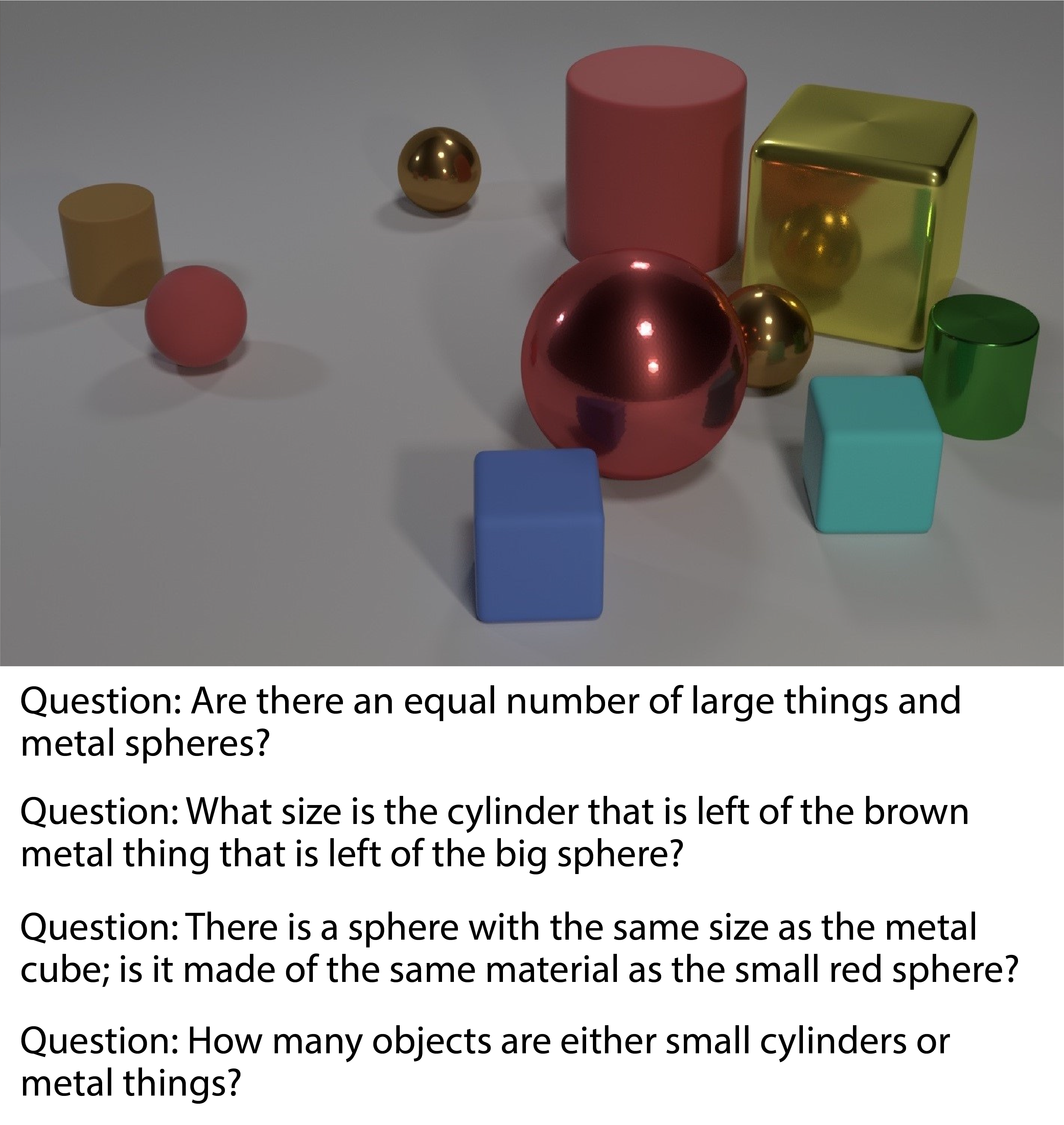}
\caption{An example from CLEVR. Figure from \cite{clevr}}
\end{figure}

A unique feature of CLEVR is that it provides structural representations of questions as extra annotation. Each question is represented by a sequence of functions which reflect the correct reasoning steps. CLEVR questions are longer and more complex than questions in general VQA datasets. This allows CLEVR to generate a greater variety of unique questions which mitigates question conditional biases. Long questions also correlate to longer reasoning steps necessary to arrive at the answer. 
\par
One drawback of CLEVR is that it contains relatively simple and basic visuals. Models can overfit due to smaller variety of objects and object features. To test this and to also test compositional generalization, the authors synthesized two new versions of CLEVR: 1) In condition A, all cubes are gray, blue, brown, or yellow and all cylinders are red, green, purple, or cyan; 2) In condition B, these shapes swap color palettes. Both conditions contain spheres of all eight colors. Good models should be able to train on one set and perform well on the other. As CLEVR does not contain real images or natural language questions, the authors suggest using it only as a diagnostic tool to test a model's compositional ability. CLEVR also has a CLEVR-humans variation where the questions are human provided natural language questions.

\subsection{Diagnostic datasets}
Diagnostic datasets are specialized in the sense that they test a model's ability in a particular area. They are usually small in size and are meant to complement larger, more general datasets by diagnosing the model's performance in a distinct area which may not have pronounced results in the more general dataset.

\begin{table*}[t]
\centering
\begin{tabular}{ |M{2cm}|M{1cm}|M{1cm}|M{1cm}|M{2cm}|M{2cm}|M{2cm}|M{2cm}| } 
\hline
Dataset & Number of Images & Number of QA pairs & Type & Image source & QA Source & Evaluation Metric & Diagnosis type \\
\hline
C-VQA\cite{cvqa} & 123K & 370K & OE and MC & Vqa-v1 & Vqa-v1 & Vqa metric & Composition \\
\hline
VQA-CP v1\cite{vqacp} & 123K & 370K & OE and MC & Vqa-v1 & Vqa-v1 & Vqa metric & Composition \\
\hline
VQA-CP v2\cite{vqacp} & 123K & 658K & OE and MC & Vqa-v2 & Vqa-v2 & Vqa metric & Composition \\
\hline
ConVQA\cite{convqa} & 108K & 6.7M & OE & VG & auto & Perfect-con, avg-con & Consistency \\
\hline
VQA-Rephrasings\cite{vqa-rephrasings} & 40K & 160K & OE & Vqa-v2 & manual & Vqa metric and consensus score & Consistency \\
\hline
VQA P2\cite{vqa-p2} & 16K & 36K & OE & Vqa-v2 & auto & Vqa metric and consensus score & Consistency \\
\hline
VQA-HAT\cite{vqa-hat} & 20K & 60K & OE & Vqa-v1 & Vqa-v1 & Rank-correlation & Visual attention \\
\hline
VQA-X\cite{vqa-x} & 28K & 33K(QA), 42K(E) & OE & Vqa-v1 and vqa-v2 & manual & BLEU-4, METEOR, ROUGE, CIDEr, SPICE and human evaluation & Textual explanation \\
\hline
VQA-E\cite{vqa-e} & 108K & 270K & OE & Vqa-v2 & Vqa-v2(QA), auto(Explanation) & BLEU-N, METEOR, ROUGE-L, and CIDEr-D & Textual  explanation \\
\hline
HowManyQA\cite{irlc} & 47K & 111K & OE & Vqa-v2 + VG & Vqa-v2 + VG & Vqa metric, RMSE(Root mean squared error) & Counting
(simple) \\
\hline
TallyQA\cite{tallyqa} & 165K & 288K & OE & Vqa-v2 + VG + TDIUC & Vqa-v2 + VG + TDIUC & Accuracy and RMSE & Counting
(simple and complex) \\
\hline
ST-VQA\cite{stvqa} & 23K & 32K & OE & Various datasets & manual & Normalized Levenshtein distance & Scene text \\
\hline
Text-VQA\cite{textvqa} & 45K & 28K & OE & Open images v3 & manual & Vqa metric & Scene text \\
\hline
\end{tabular}
\caption{Statistics of diagnostic VQA datasets.}
\end{table*}

\subsubsection{C-VQA and VQA-CP}
C-VQA\cite{cvqa} is a rearranged version of VQA-v1. It is meant to help evaluate a model's compositional ability, i.e., the ability to answer questions about \textit{unseen compositions of seen concepts}.  For instance, a model is said to be compositional if it can correctly answer [“What color are the horses”, “white”] without seeing this QA pair during training, but perhaps having seen [“What color are the horses?”, “red”] and [“What color are the dogs?”, “white”]. C-VQA is created by rearranging the train and val splits of VQA-v1 so that the QA pairs in C-VQA test split are compositionally novel with respect to those in C-VQA train split. For instance, “tennis” is the most frequent answer for the question type “what sport” in C-VQA train split whereas “skiing” is the most frequent answer for the same question type in C-VQA test split. However for VQA-v1, “tennis” is the most frequent answer for both the train and val splits. Like C-VQA, VQA-CP\cite{vqacp} reorganizes VQA-v1 and VQA-v2 so that the answer distribution of each question type is different in the train and test splits.

\begin{figure}%
    \centering
    \subfloat[\centering]{{\includegraphics[width=4cm, height=3.5cm]{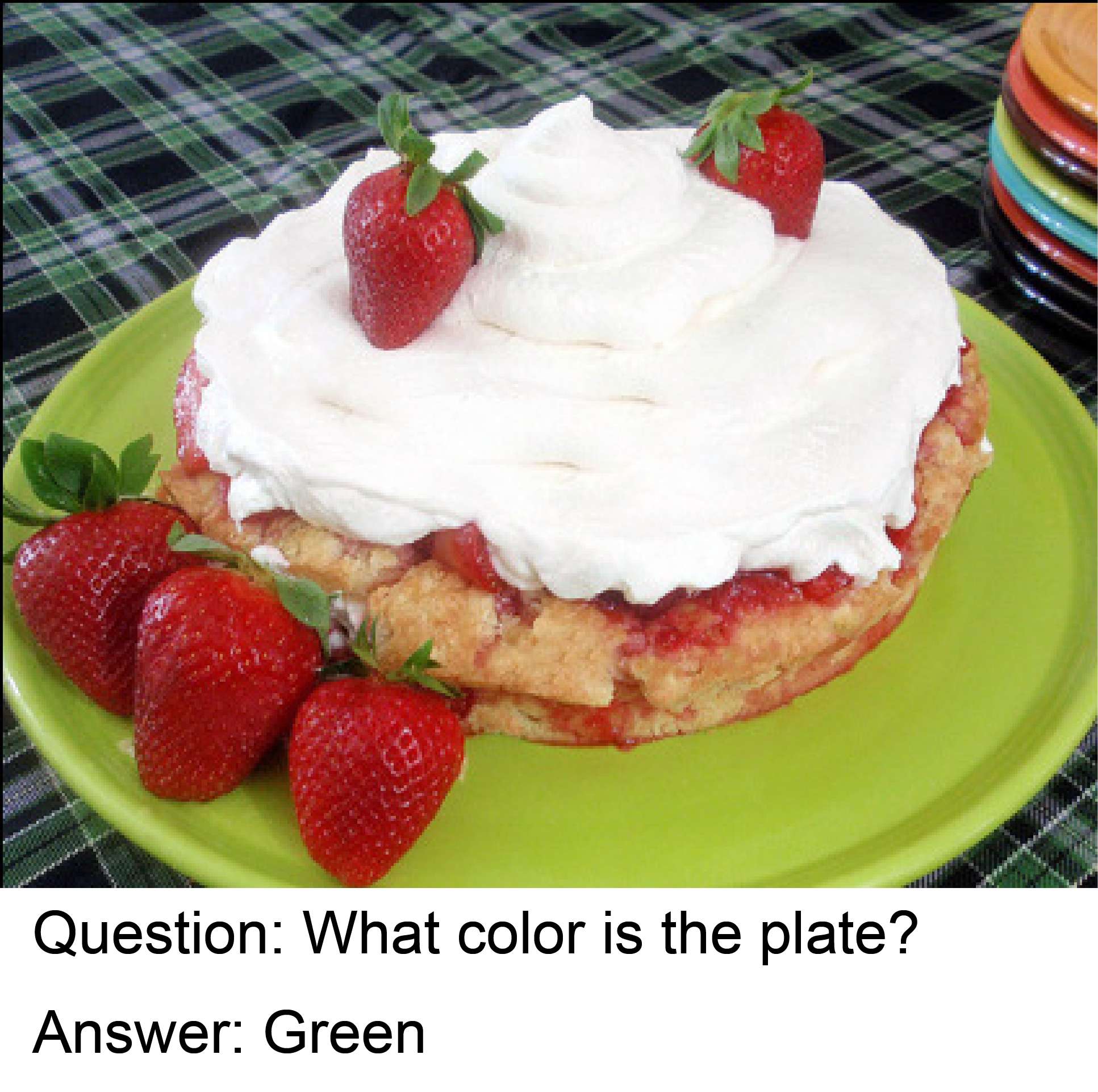} }}%
    \quad
    \subfloat[\centering]{{\includegraphics[width=4cm, height=3.5cm]{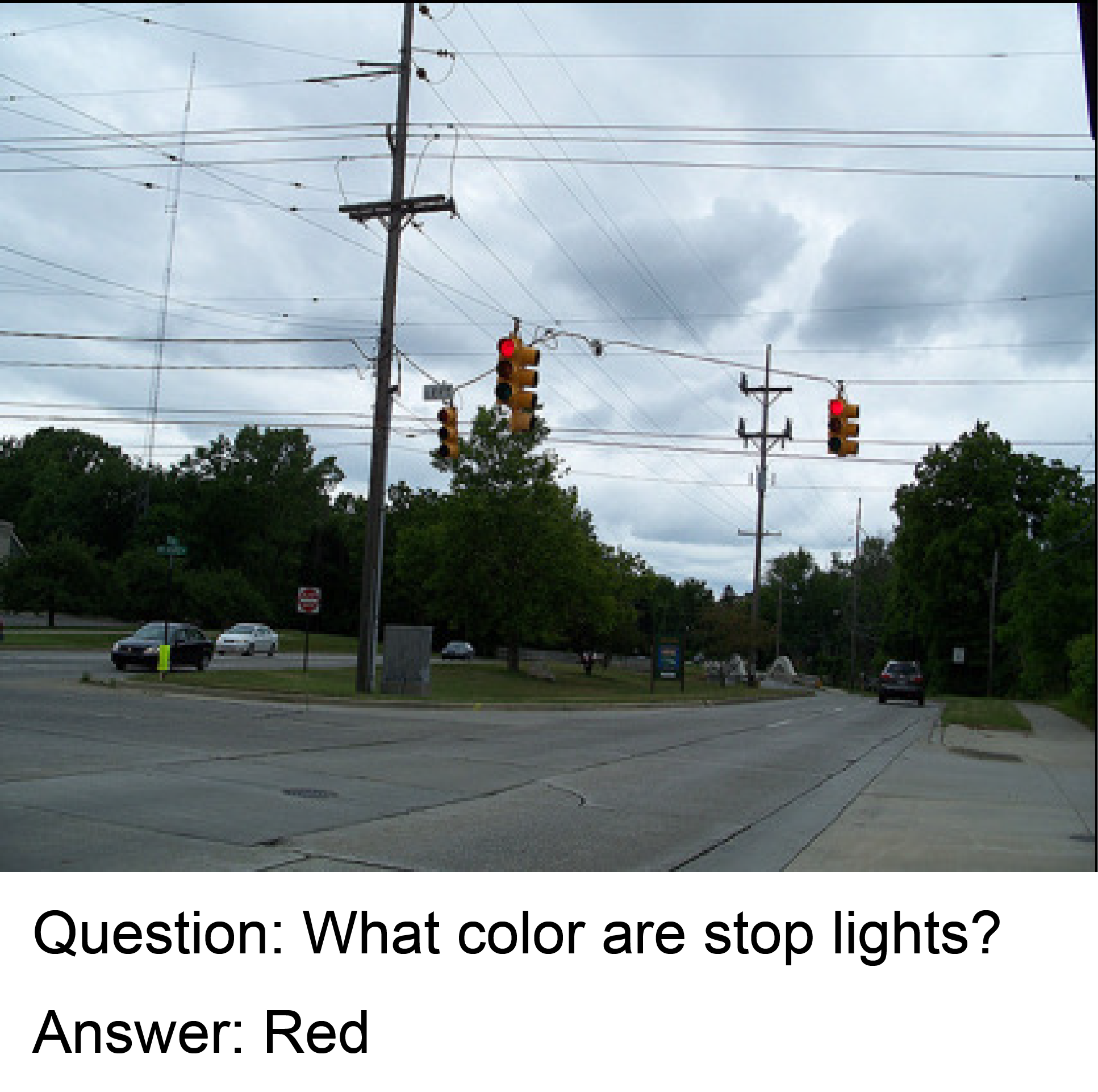} }}%
    \subcaption*{Training}\vspace{.5cm}
    \subfloat[\centering]{{\includegraphics[width=4cm, height=3.5cm]{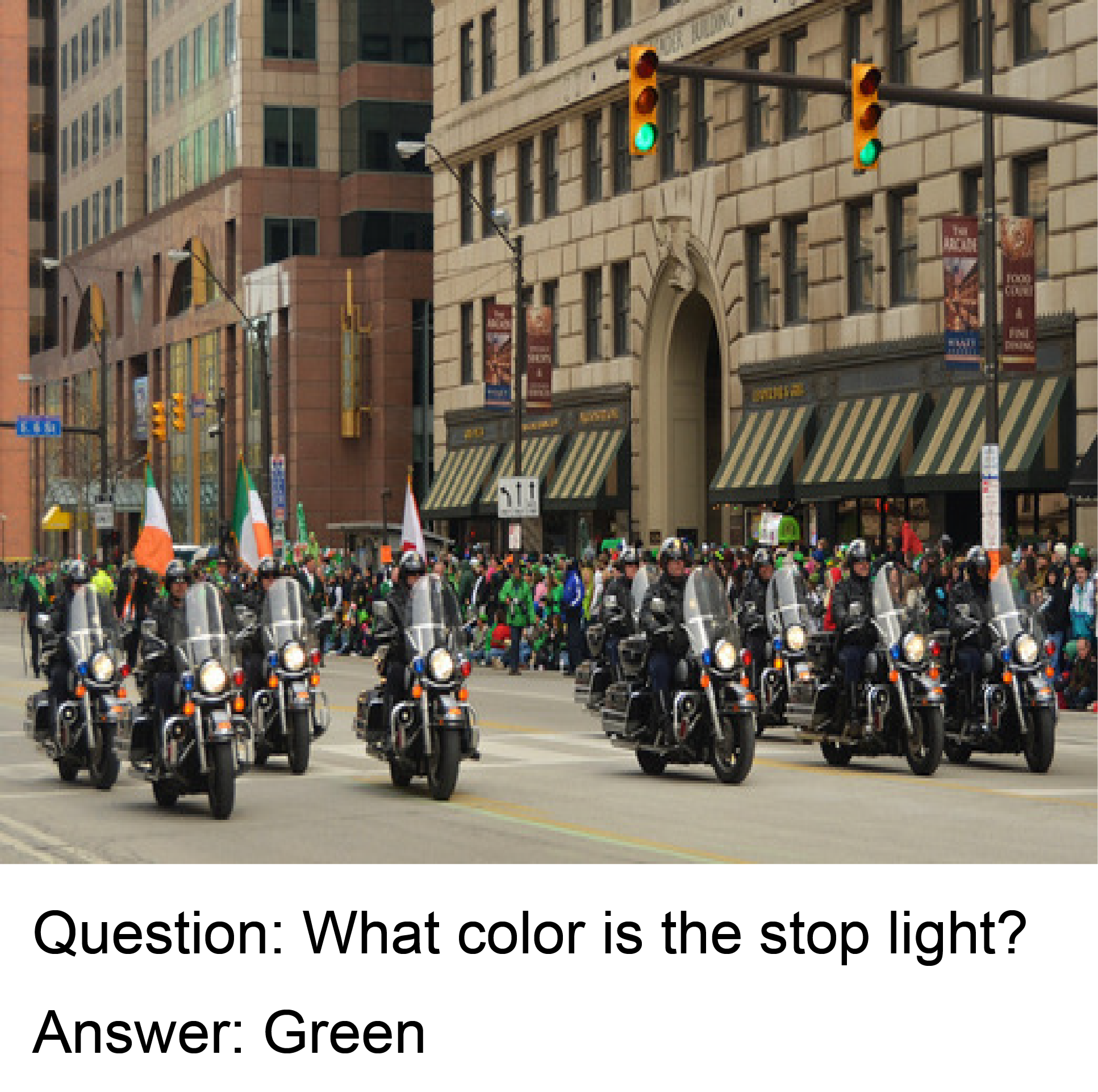} }}%
    \quad
    \subfloat[\centering]{{\includegraphics[width=4cm, height=3.5cm]{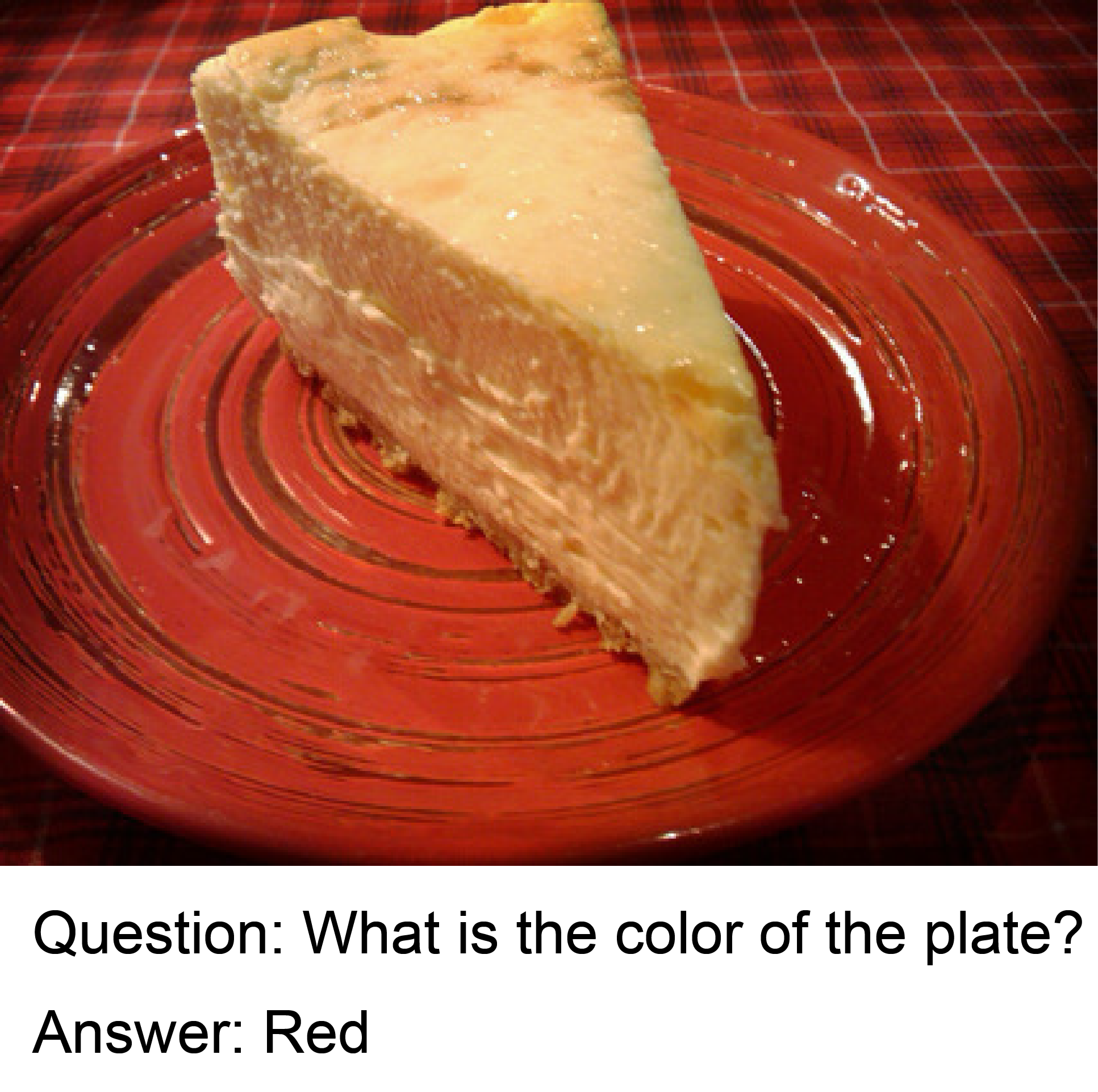} }}%
    \subcaption*{Testing}
    \caption{An example from CVQA. Figure from \cite{cvqa}}%
\end{figure}

\subsubsection{VQA-E, VQA-X and VQA-HAT}
VQA-E\cite{vqa-e} and VQA-X\cite{vqa-x} argue that a VQA model should provide an explanation along with the answer so that it can be verified that the answer was deduced properly without exploiting dataset biases. To facilitate this, both datasets provide explanations along with the usual image-question-answer triplet in its training set. This explanation annotation can teach a model to produce explanations for unseen instances. VQA-E contains 270K textual explanations for 270K QA pairs and 108K images while VQA-X is smaller containing 42K explanations for 33K QA pairs and 28K images. Explanations in VQA-X are human-provided while in VQA-E they are automatically generated. VQA-X also provides visual explanations in the form of human-annotated attention maps for its images. VQA-HAT\cite{vqahat} is another dataset that provides attention annotations. To collect data for VQA-HAT, human subjects were presented with a blurred image and then were asked a question about that image. They were instructed to deblur regions in the image that would help them to answer the question correctly. The regions they chose to deblur are thought to correspond to the regions humans naturally choose to focus on in order to deduce the answer.

\subsubsection{VQA-Rephrasings}
VQA-Rephrasings\cite{vqa-rephrasings} shows that most state-of-the-art VQA models are notoriously brittle to linguistic variations in questions. VQA-Rephrasings provides 3 human-provided rephrasings for 40k questions spanning 40k images from the VQA-v2 validation dataset. They also provide a model-agnostic method to improve the ability of answering rephrased questions.

\begin{figure}[ht]%
    \centering
    \subfloat[\centering]{{\includegraphics[width=4cm, height=5.5cm]{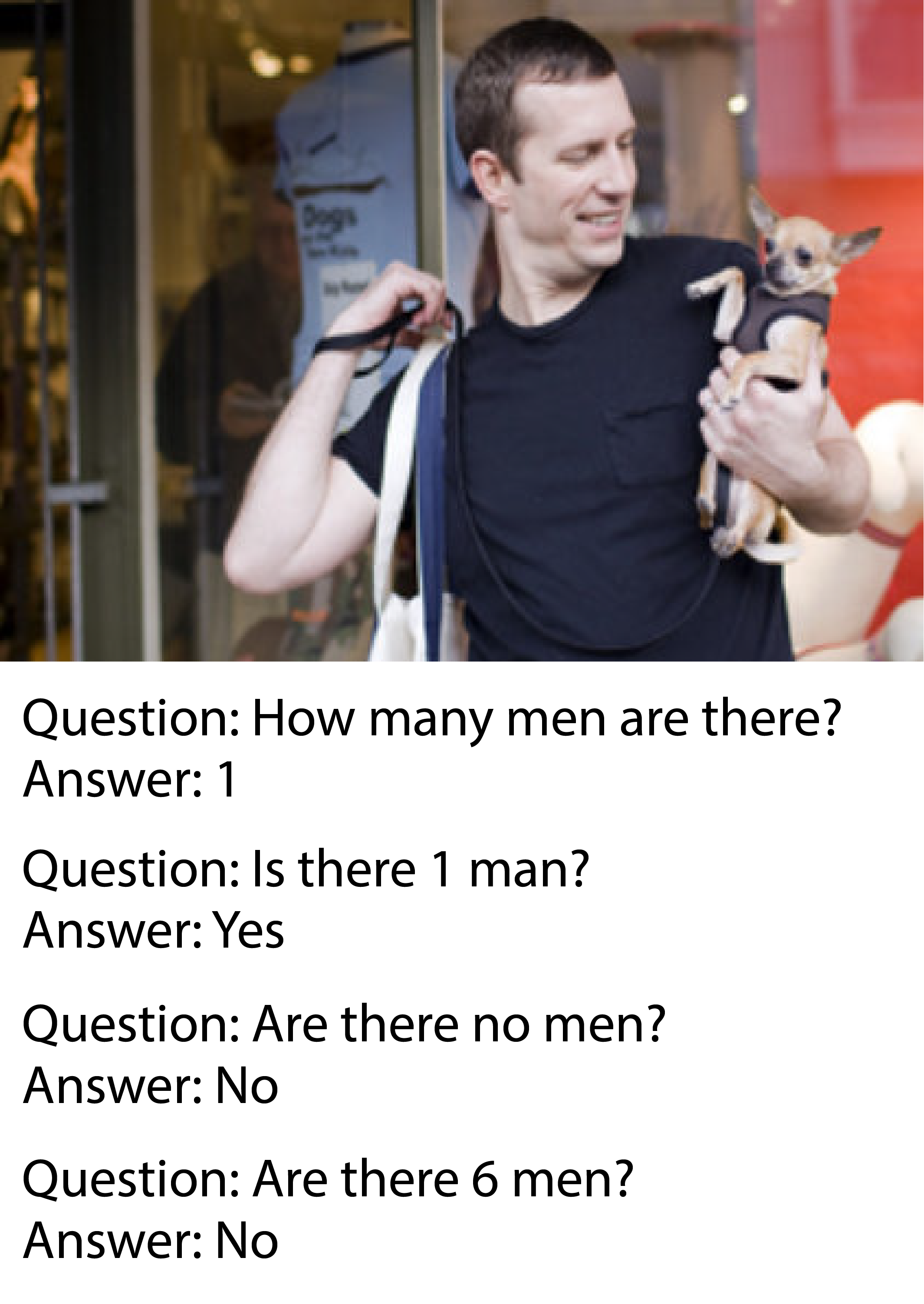} }}%
    \quad
    \subfloat[\centering]{{\includegraphics[width=4cm, height=5.5cm]{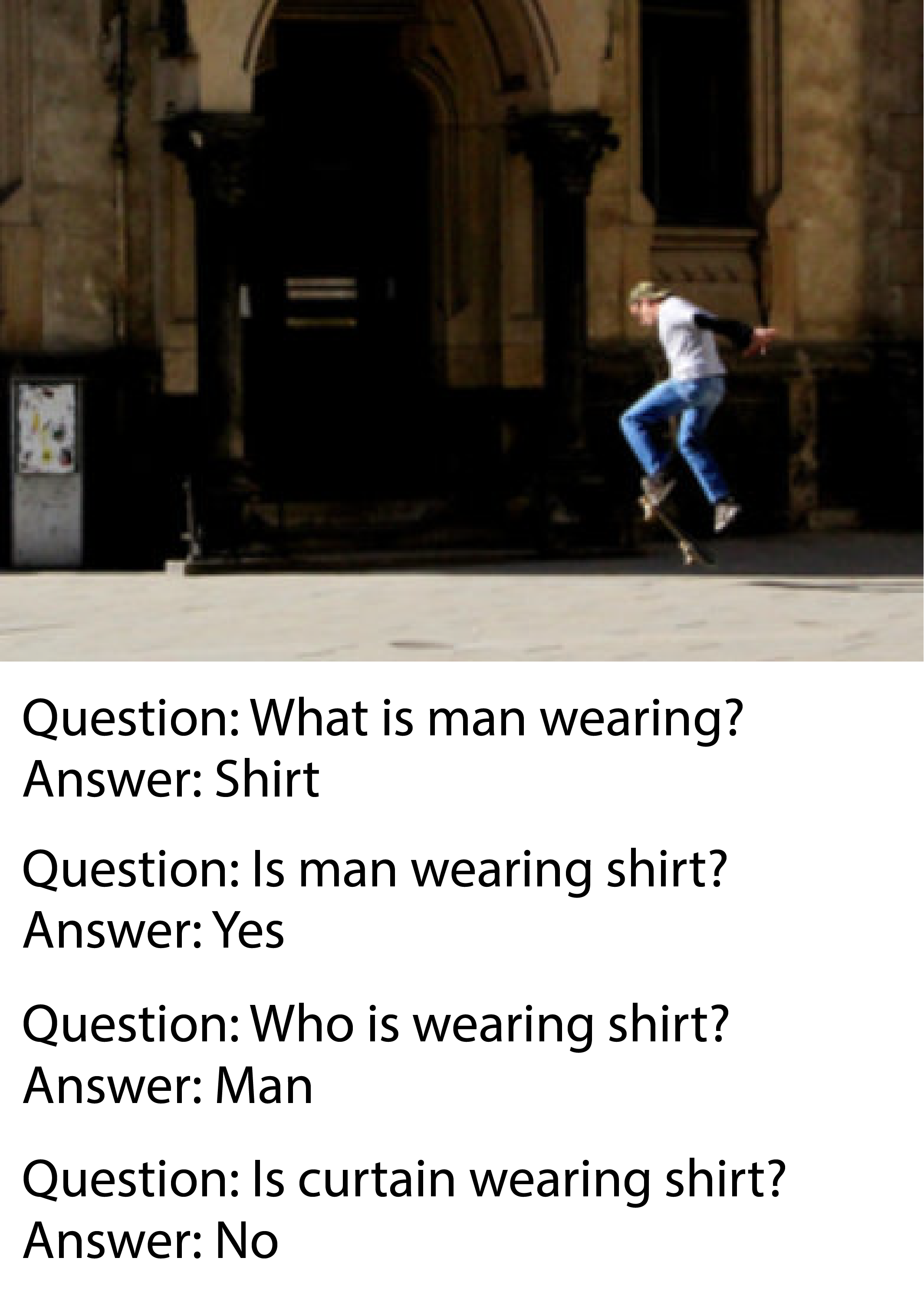} }}%
    \caption{An example from ConVQA. Figure from \cite{convqa}}%
\end{figure}

\subsubsection{ConVQA}
ConVQA\cite{convqa} introduces the task of belief consistency in VQA models, i.e., the task of consistently answering questions that differ in linguistics but share the same semantics. For example, if the answer to “Is it a vegetarian pizza?” is “yes”, then the answer to “Is there pepperoni on the pizza?” should be “no”. Previous VQA datasets did not have such multiple questions about a single concept to test the consistency of a model. There are a total of 6,792,068 QA pairs in 1,530,979 consistent sets over 108,077 images in ConVQA.

\begin{table*}[ht]
\centering
\begin{tabular}{ |M{1cm}|M{1cm}|M{1cm}|M{0.5cm}|M{1cm}|M{2cm}|M{1cm}|M{2cm}|M{2cm}|M{2cm}| } 
\hline
Dataset & Number of Images & Number of QA pairs & Type & Question types & Image source & QA source & Metric & External knowledge annotation & Knowledge base used in construction \\
\hline
FVQA\cite{fvqa} & 2190 & 5826 & OE & 32 & MS COCO and imagenet & Manual & Exact matching, WUPS & Supporting fact
(arg1, rel, arg2) & Combination of DBpedia, WebChild, and ConceptNet \\
\hline
KB-VQA\cite{kvqa} & 700 & 2402 & OE & 23 & MS COCO & Mixed & Exact matching, WUPS, Manual & none & DBpedia \\
\hline
OK-VQA\cite{okvqa} & 14K & 14K & OE & 10 & MS COCO & Manual & Vqa metric & none & none \\
\hline
\end{tabular}
\caption{Statistics of KB datasets.}
\end{table*}

\subsubsection{HowManyQA and TallyQA}
HowManyQA (introduced in \cite{irlc}) was made by combining counting questions from VQA-v2 and Visual Genome resulting in 106,356 counting questions. TallyQA\cite{tallyqa} points out that most counting questions in VQA datasets and HowManyQA are simple which require nothing more than object detection such as, “How many dogs are there?”. They present a dataset which contains a large number of complex counting questions, for example, “How many dogs are laying on the grass?”. These questions require more complex reasoning about objects, their attributes and their relationships with other objects and background regions. TallyQA contains 287,907 questions of which 76,477 are complex questions.

\subsubsection{Text-VQA and ST-VQA}

\begin{figure}[ht]%
    \centering
    \subfloat[\centering]{{\includegraphics[width=4cm, height=4cm]{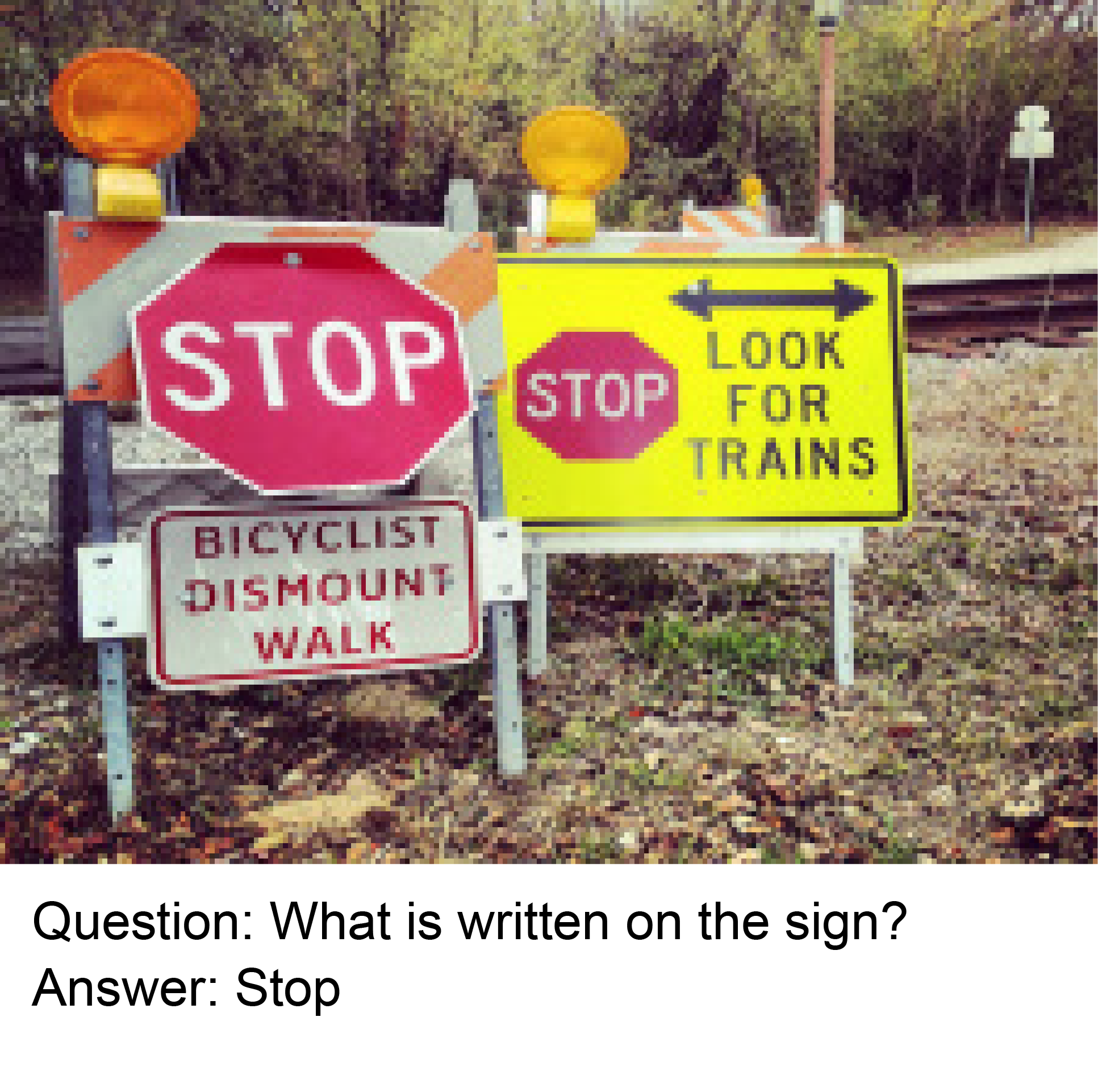} }}
    \subfloat[\centering]{{\includegraphics[width=4cm, height=4cm]{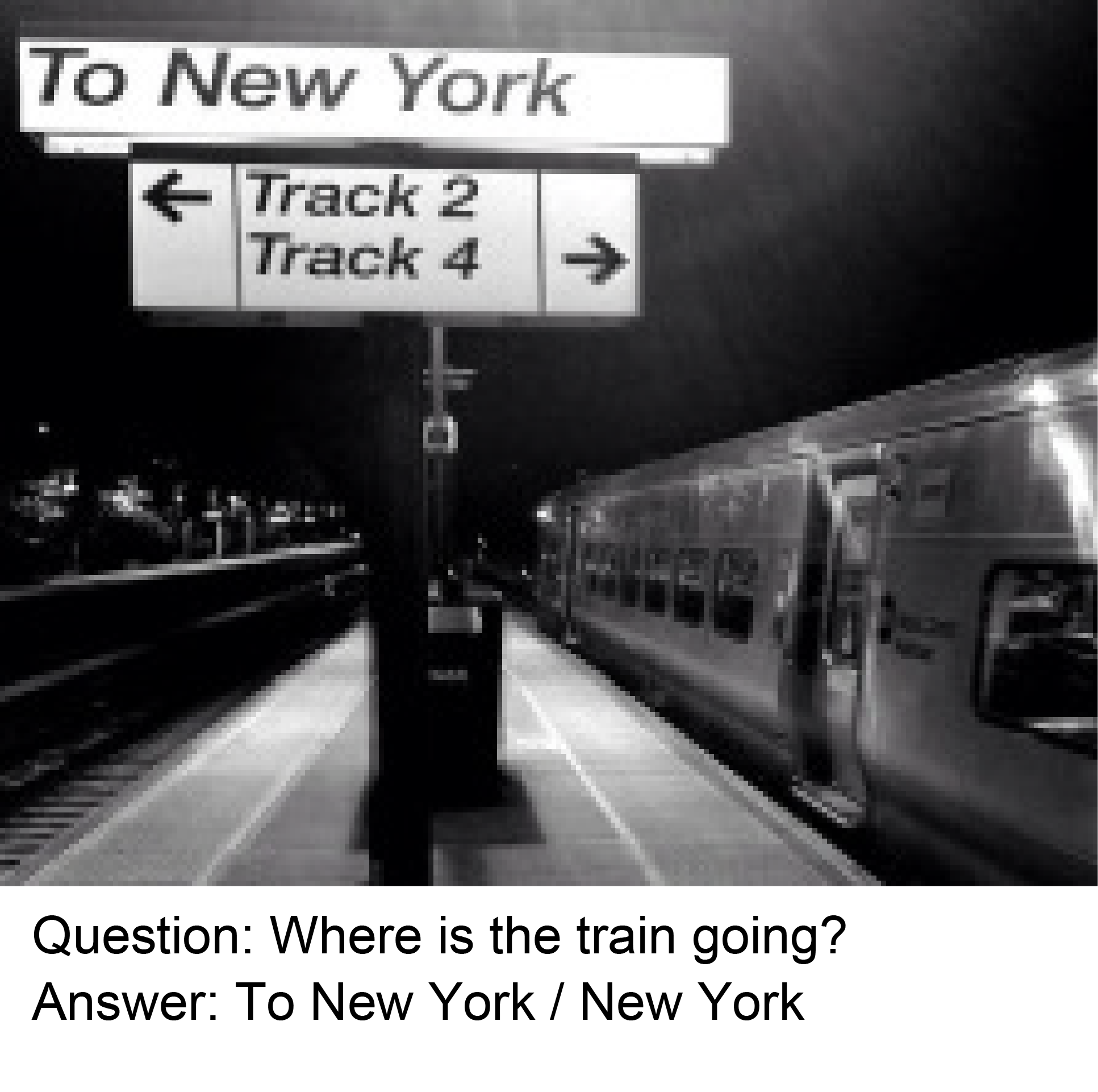} }}%
    \quad
    \caption{An example from ST-VQA. Figure from \cite{stvqa}}%
\end{figure}

A lot of questions humans ask about an image tend to be about text in the image, for example, “What is the number on the license plate?”. Such types of questions are rare in general VQA datasets. TextVQA\cite{textvqa} contains 45,336 questions about 28,408 images that require reasoning involving the text present in the image. Each image-question pair has 10 ground truth answers provided by humans. ST-VQA\cite{stvqa} contains 31,791  scene text related questions about 23,038 images. ST-VQA ensures that all questions are unambiguous and all images contain a minimum amount of text. It ensures bias is not present in question and answer distribution. It also proposes a new evaluation metric more suited for this particular task.

\subsection{KB datasets}
Sometimes it is not possible to answer a question with only the information present in the image. In such cases, the required knowledge has to be acquired from external sources. This is where KB datasets come in. They provide questions that require finding and using external knowledge. KB datasets can teach a model to know when it needs to search for absent knowledge and how to acquire that knowledge. Even the largest VQA datasets can not contain all real-world concepts. So VQA models should know how to acquire knowledge about unseen concepts to achieve true generalizability. It should be noted that KB datasets are different from EKBs or External Knowledge Bases which are large structured databases compiling various world information. Examples of such EKBs are, large scale KBs constructed by human annotation, e.g., DBpedia\cite{dbpedia}, Freebase\cite{freebase}, Wikidata\cite{wikidata} and automatic extraction from unstructured/semi-structured data, e.g., YAGO\cite{yago2},\cite{yago3}, OpenIE\cite{openie1},\cite{openie2},\cite{openie3}, NELL\cite{nell}, NEIL\cite{neil}, WebChild\cite{webchild}, ConceptNet\cite{conceptnet}.

\subsubsection{FVQA}
FVQA\cite{fvqa} is a dataset where most of the questions(5826 questions about 2190 images) require external information to answer. To support this, FVQA adds an additional “supporting fact” to each image-question-answer triplet. Each supporting-fact is represented as a structural triplet, such as (Cat, CapableOf, ClimbingTrees), which can be used to answer the question. These facts mimic common-sense and external knowledge that help humans(as such should help VQA models too) answer questions that require outside knowledge. By using these facts, FVQA teaches a model to look for the appropriate supporting fact from a facts-database and use that to deduce the answer.

\begin{figure}[ht]
\centering
\includegraphics[width=10cm,
  height=6cm,
  keepaspectratio]{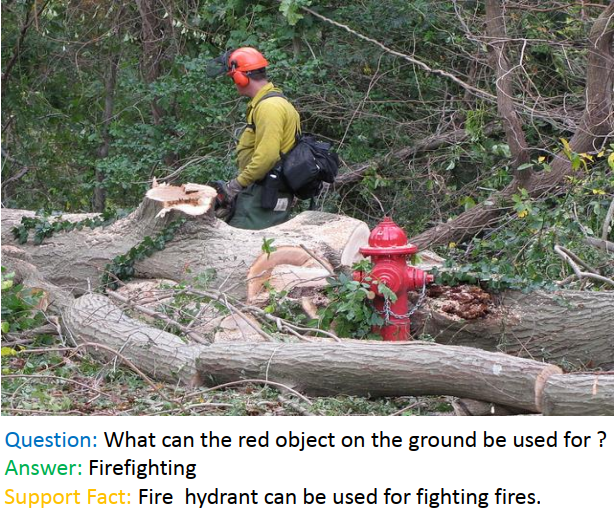}
\caption{An example from FVQA. Figure from \cite{fvqa}}
\end{figure}

\subsubsection{KB-VQA}
KB-VQA\cite{kvqa} is a small dataset containing 2402 questions about 700 images based on 23 question templates. Among these, 1256 are visual questions(does not require external information), 883 require common-sense knowledge and 263 require querying an external knowledge base.

\subsubsection{OK-VQA}
OK-VQA\cite{okvqa} contains 14,055 questions about 14,031 images. All questions were collected so that they require outside knowledge to answer. OK-VQA is larger than both FVQA and KB-VQA. Care was taken so that various biases such as, bias in answer distribution were reduced as much as possible.

\subsection{Evaluation}
A model's performance being correctly evaluated depends on the evaluation metric used. Unfortunately, a major problem of VQA is that there is no widely agreed upon evaluation metric. Many different metrics have been proposed.

\subsubsection{Exact matching}
The simplest metric where predictions are considered correct if they match the answer exactly. There are obvious problems with this approach. Many questions have multiple valid answers. The number of valid answers increases with the number of words. This metric can be used for datasets where most of the answers are single-word. But for datasets like Visual Genome where 27\% of answers contain three or more words, this metric performs poorly.

\subsubsection{WUPS}
WUPS or Wu-Palmer similarity test\cite{wups} is a metric where the predicted answer is given a rating between $0$ and $1$ with respect to the ground-truth answer. The ranking is calculated by tracing their common subsequence in a taxonomy tree and it reflects how “semantically” similar the two words are. For example, “bird” and “eagle” would have a rating closer to $1$ whereas “bird” and “dog” would have a lower rating. But some word pairs are ranked high in this scheme even though they are very loosely related. Moreover, some questions are such that only one answer is correct whereas other answers, despite being semantically similar, are wrong. For example, a question with “What color ..” can only have one correct answer, e.g., “red” while other colors like “black” are wrong answers.

\subsubsection{Consensus}
This was the metric used for the VQA-v1 dataset. The metric is,
\begin{equation}
    Accuracy_{VQA} = min(\frac{n}{3},1)
\end{equation}
where $n$ is the number of annotators with the same answer as the predicted one. Basically, this metric considers any prediction correct if it matches with the answers of at least 3 annotators. This scheme allows multiple valid answers. There are some problems with this metric. It turns out that not all questions have annotators agreeing what the answer should be. Especially 59\% of “why” questions have no answer with more than two annotators which means the metric will consider any answer for these questions incorrect. Moreover, there are questions where directly conflicting answers, like “yes” and “no”, have at least three annotators. This means that both “yes” and “no” are considered valid answers for these questions.

\subsubsection{TDIUC metrics}
Because of skewed question and answer distribution, normal evaluation metrics do not capture how well a model is performing across all question and answer categories. TDIUC\cite{tdiuc} introduces A-MPT(Arithmetic mean per type) and H-MPT(Harmonic mean per type). A-MPT evaluates average accuracy like previous metrics. H-MPT measures the ability of a system to have high scores across all question-types and is skewed towards the lowest performing categories. TDIUC also introduces N-MPT(normalized MPT) to compensate for bias in the distribution of answers within each question-type. This is done by computing the accuracy for each unique answer separately within a question-type and then averaging them together. A large discrepancy between unnormalized and normalized scores suggests that the model is not generalizing to rarer answers.

\subsubsection{GQA metrics}
GQA\cite{gqa} introduces 5 new metrics for detailed analysis of the true capability of a model. \textit{Consistency} checks that the model provides consistent answers for different questions based on the same semantics. For example, If the model answers the question “What is the color of the apple to the right of the plate?” correctly with red, then it should also be able to infer the answer to the question “Is the plate to the left of the red apple?”. \textit{Validity} checks whether a given answer is in the question scope, e.g., responding some color, and not some size, to a color question. \textit{Plausibility} score goes a step further measuring whether the answer makes sense given the question. For example, red and green are plausible apple colors and conversely purple is implausible. \textit{Distribution} measures the overall match between the true answer distribution and the predicted distribution. It allows us to see if the model has generalized to rarer answers. For attention-based models, \textit{Grounding} checks whether the model attends to regions within the image that are relevant to the question. This metric evaluates the degree to which the model grounds its reasoning in the image rather than just making educated guesses.

\begin{figure*}[ht]
\centering
\includegraphics[width=18cm,
  height=8cm,
  keepaspectratio]{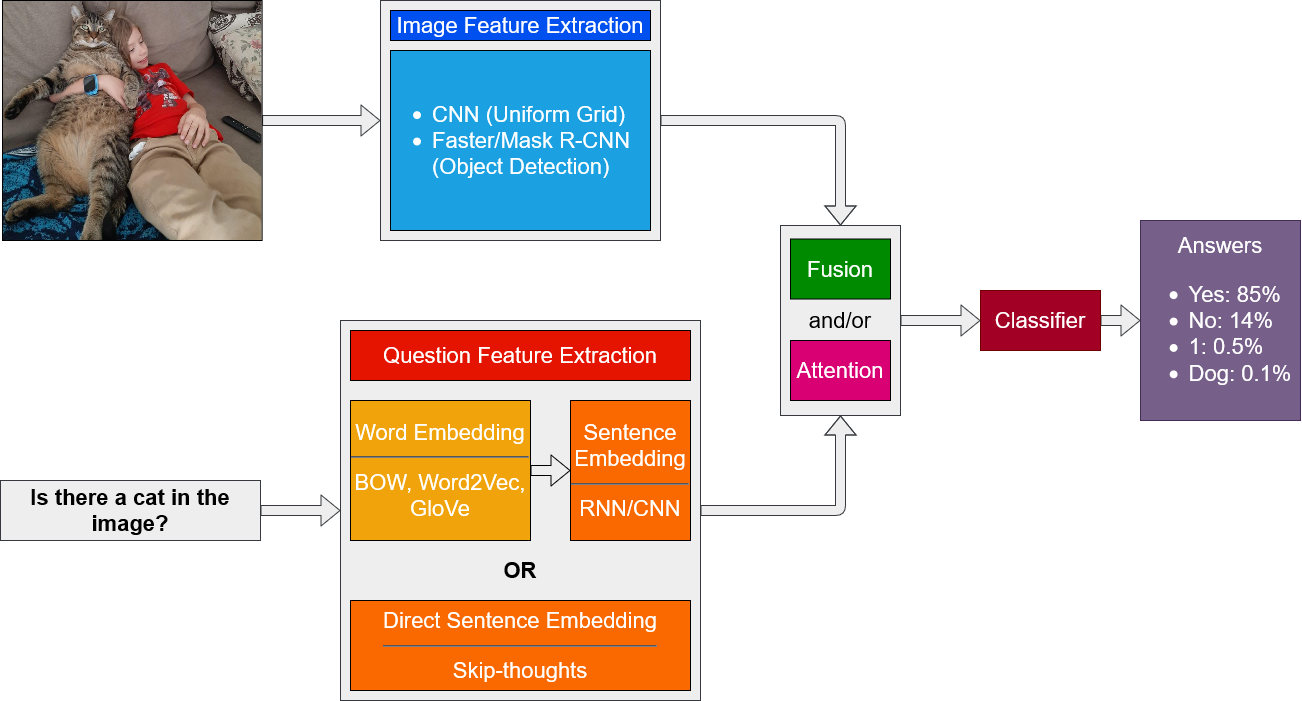}
\caption{Basic approach of VQA algorithms}
\end{figure*}

\section{Algorithms}
Most VQA algorithms follow a basic structure:
\begin{itemize}
\item
Image representation
\item
Question representation
\item
Fusion and/or Attention
\item
Answering
\end{itemize}

\subsection{Image representation}

Image representation in VQA is done using either pre-trained CNNs(Convolutional Neural Network) or object detectors. Some examples of pre-trained CNNs that have been used are GoogLeNet\cite{googlenet}, VGGNet\cite{vggnet}, ResNet\cite{resnet}. Among these, ResNet seems to consistently give higher accuracies than other CNNs. For detecting object-regions, the primary choice has been Faster R-CNN\cite{fasterr-cnn}. Mask R-CNN\cite{maskr-cnn} is a more recent detector that seems to be more robust and better at detection than Faster R-CNN. \par

Pre-trained CNNs like ResNet have been pre-trained on millions of images from large databases like ImageNet\cite{imagenet}. Images from these datasets were also used to construct most general VQA datasets which makes transfer learning possible. When given an input image, a CNN goes through several convolution and pooling layers to produce a $C\times W\times H$ shaped output. Here, $W$ and $H$ refers to the modified width and height of the image and $C$ refers to the number of convolution filters. Each convolution filter can be thought to detect a certain pattern. \par

Object detectors work a bit differently. They produce multiple bounding boxes. Each bounding box usually contains an object belonging to a specific object class. Here “object” can mean things like sky, cloud, water which aren't considered objects in the traditional sense. Faster R-CNN produces rectangular bounding boxes which means an object doesn't always fit inside its box. Object detection also suffers from overlapping issues. The same object can be part of multiple bounding boxes all overlapping each other. For example, Consider an image containing a person wearing a shirt. The shirt can be part of two overlapping bounding boxes, the bounding box containing 'shirt' and the bounding box containing 'person'. Bounding boxes of different objects can also overlap each other especially if one object occludes another object behind it. Mask R-CNN\cite{maskr-cnn} produces polygonal masks which reduces these problems. Most VQA algorithms using object detectors typically use the top 24-32 predicted objects from the image.\par

\begin{figure}[ht]
\centering
\includegraphics[width=15cm,
  height=6cm,
  keepaspectratio]{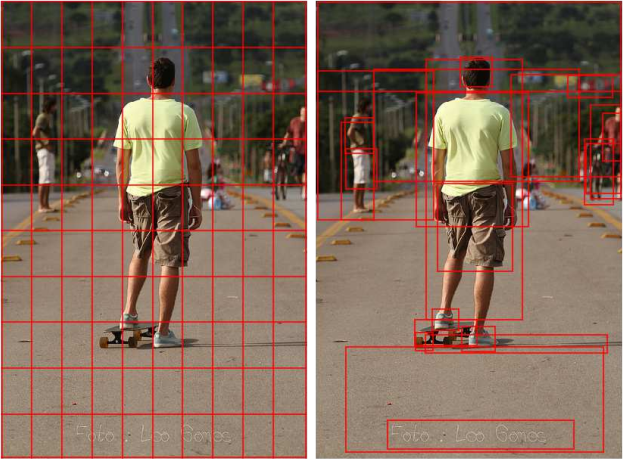}
\caption{CNN(left) and Faster R-CNN(right). Figure from \cite{butd}}
\end{figure}

\begin{figure}[ht]
\centering
\includegraphics[width=10cm,
  height=4cm,
  keepaspectratio]{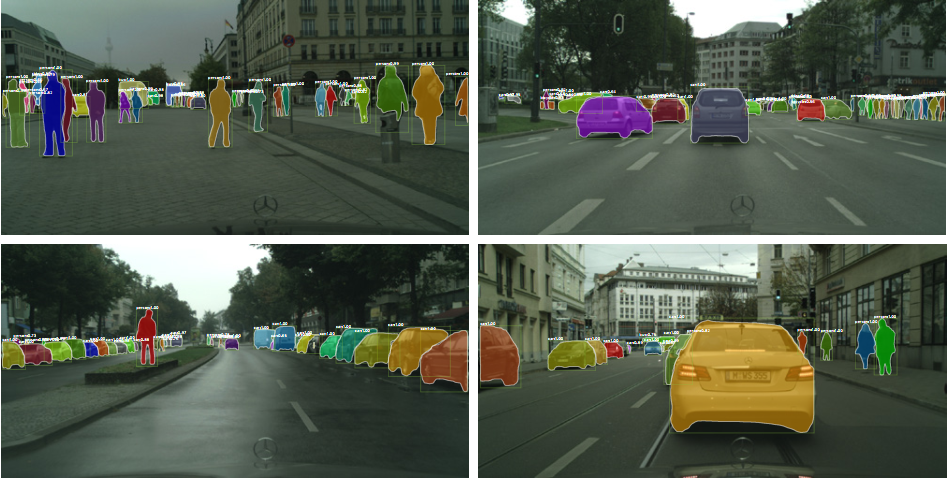}
\caption{Mask R-CNN. Figure from \cite{maskr-cnn}}
\end{figure}

According to the method used, image representation can be divided into \textit{grid-based} and \textit{object-based} approaches. CNNs divide any image into a uniform grid(typically $14\times 14$) and object detectors divide the image into multiple object-regions(typically 32). A uniform grid structure means sometimes an object can overlap into several adjacent cells. Irrelevant noise can be an issue as the whole image is used but this can be mitigated through attention mechanisms. Object detectors break the image down into salient objects which may provide a stronger training signal as the scene is organised into meaningful units. But they ignore all area outside of the predicted bounding boxes which may result in information loss. Visual information about broader semantics such as scene and sentiment is also lost. 
\par
Which method of visual representation is ideal is a topic of major debate in VQA. Earlier models universally used pre-trained CNN features. BUTD\cite{butd} was the  first major VQA model to utilize object detection. Since BUTD, most models have opted to use Faster R-CNN pre-trained on the Visual Genome object detection task. But recently \cite{grid-defense} have presented findings arguing for using grid features. Like BUTD, \cite{grid-defense} also pre-trains Faster R-CNN for VG object detection but unlike BUTD, they discard region prediction during the inference step and directly use the grid features from the last convolution layer. They identify that pre-training for object detection is an important step that incorporates richer visual understanding but extracting object features are not essential for answering visual questions. \par

\subsection{Question representation}

Question representation in VQA is usually done by first embedding individual words and then using an RNN or a CNN to produce an embedding of the entire question. \par

Some earlier models used one-hot vectors and BOW(Bag of words) to embed individual words. Then they either used them directly or fed them to an RNN or CNN. Later models used a pre-trained word embedding model such as: Word2Vec\cite{word2vec} or GloVe\cite{glove}. Word2Vec was built using skip-gram and negative sampling. It produces similar vectors for words with similar semantic meanings. GloVe uses word-word co-occurrence probability on the basis that words that occur frequently together have a higher probability of being semantically connected. More recent works have utilized language models such as, ELMo\cite{elmo}, BERT\cite{bert}, and GPT\cite{gpt-1}\cite{gpt-2}\cite{gpt-3} which have set new state-of-the-art performances across a range of NLP tasks. These models are pre-trained on a huge amount of natural language data and can be expected to produce sufficiently useful representations for the VQA task. \par

After getting the word embeddings, they are usually passed to an RNN such as, GRU\cite{gru}, LSTM\cite{lstm}, bi-LSTM, and bi-GRU. Another option is to also use a CNN for question representation. Some models skip the word embedding and directly embed the whole question using a sentence embedding model like Skip-thought\cite{skip-thought}. \par

In terms of which RNN model works best for VQA, there has not been any conclusive results favoring any particular model. Another question to ask is “Does RNN work better than CNN?”. Although RNNs usually outperform CNN in many NLP related tasks, the same can not be definitely said for VQA. \cite{cnnbetter} performs a detailed comparison between RNN and CNN for VQA. They conclude that CNN is better suited for VQA than RNN because of the limited length of question sentences which prevents RNNs from using their full potential. Understanding a question properly requires identifying keywords which CNN does better as it treats individual words as features. We think that comparison between CNN and RNN for VQA is something that should be looked into further.

\subsection{Fusion/Attention}

For the network to predict answers, it has to jointly reason over both question and image. The interaction of the visual and textual domain in VQA is either done directly through multimodal fusion or indirectly through attention mechanisms. We will talk about both methods in the upcoming sections.

\subsection{Answering}

For open-ended VQA, answers are predicted either through a generative approach or a non-generative approach. In the non-generative approach, all or the most frequent unique answers in the dataset are set as individual classes. This approach is easier to evaluate and implement but has the obvious shortcoming of not being able to predict answers not seen during training. Another option is to generate the answer word by word using an RNN but there are no good methods of evaluating the generated answers which means almost all models adopt the non-generative approach. For multiple-choice VQA, answer prediction is treated as a ranking problem where each question, image and answer trio gets a score and the answer with the highest score is selected.
\par
There are two types of non-generative approach, single-label classification and multi-label regression. In single-label classification, models output a probability distribution over the answer classes using softmax. In this approach, models are trained to maximise probability for a single correct answer, usually the ground truth answer most annotators agree on. It can be argued that this single-label approach results in a poorer training signal and a multi-label regressive approach fares better. Most VQA questions have multiple correct answers, so it makes sense to produce multiple labels. The most popular VQA dataset VQA-v1 actually provides multiple ground truth answers. The answers are assigned soft accuracies according to the metric, $min(\frac{\text{\# humans that provided that answer}}{3},1)$. This makes evaluating multiple correct answers possible as answers with fewer annotators still get a partial score. 
\par
For a long time, VQA models used single-label classification. BUTD\cite{butd} was the first model to use the answer scores from VQA-v1 as soft targets and cast the answering task as a regression of scores for candidate answers instead of traditional classification. BUTD argues that this method is much more effective than simple softmax classification because it allows for producing multiple answers and models the distribution of human provided annotations better. Following BUTD, most recent models use the multi-label regression approach.
\par
Almost all models represent answers as one-hot vectors for easier implementation. But this handicaps semantic understanding as there is now a gap between how the model interprets the question and how it interprets the answer, despite both coming from the same semantic space. For example, for a question with the correct answer of “dog”, most models will penalize “cat” and “german shephard” equally even though “german shephard” comes closer to the correct answer. Some works have tried to close this gap by projecting the answers into the same semantic space as the questions. Just like the questions, the answers are turned into vectors and the task of answering becomes regressing the answer vector. This way the model can have a more accurate view of the answer space where “german shephard” is closer to “dog” than “cat” resulting in a better training signal and a more nuanced understanding.

\section{Multimodal Fusion}

In order to perform joint reasoning on a QA pair, information from the two modalities have to mix and interact. This can be achieved by multimodal fusion. We divide fusion in VQA into two types, vector operation and bilinear pooling.

\subsection{Vector Operation}
Early VQA models used vector addition, multiplication and concatenation to perform fusion. They are relatively simple in terms of complexity and computational cost. In vector addition and multiplication, question and image features are projected linearly through fully-connected layers to match their dimensions. Although vector operations are simple to implement, they do not result in good accuracy. Among the three vector operations, concatenation usually has the worst performance while multiplication seems to fare the best.

\subsection{Bilinear Pooling}
Bilinear pooling\cite{bilinear} computes the outer product of the question and image feature vectors. The outer product of two vectors gives a complete representation of all possible interactions between the elements of the vectors. If $z_i$ is the $i$-th output(with bias terms omitted) then,
\begin{equation}
    z_i = x^TW_iy
\end{equation} 
where $x \epsilon \mathbb{R}^m $, $y \epsilon \mathbb{R}^n $, $z \epsilon \mathbb{R}^o $ and $W \epsilon \mathbb{R}^{m \times n \times o} $. But in VQA, directly computing the outer product is infeasible. If both image and question feature vectors have dimensions of 2048 and there are 3000 answer classes, we would need to compute 12.5 billion parameters \cite{mcb}. This is computationally expensive and may lead to overfitting. The models we discuss try to control the number of parameters by imposing various restrictions. Ultimately, we will see that these models are just different approaches towards the trade-off between expressibility and complexity or trainability. 
\par
It has been proven in \cite{countsketchproof} that the count sketch of the outer product of two vectors can be expressed as the convolution of count sketches of the two vectors,
\begin{equation}
\Psi(x \otimes y, h, s) = \Psi(x, h, s) * \Psi(y, h, s)
\end{equation}
Here, $\Psi$ is the count-sketch, $\otimes$ is the outer-product, $*$ is the convolution operator, and $h$ and $s$ are randomly sampled parameters by the model. MCB\cite{mcb} utilizes the above approach but replaces the convolution with a more efficient element-wise product in FFT(Fast Fourier Transform) space. In this way, MCB indirectly computes the outer product. 

\par
MLB\cite{mlb} argues that MCB still has too many parameters. MCB still needs to maintain a large number of parameters to overcome the bias that results from $h$ and $s$ being fixed. MLB tries to mitigate this by decomposing $W$ into $W = UV^T$ which results in,
\begin{equation}
    z_i = 1^T(U_i^Tx \circ V_i^Ty)
\end{equation}
where $\circ$ is the hadamard product. Here $U_i \epsilon \mathbb{R}^{m \times k} $ and $V_i \epsilon \mathbb{R}^{n \times k} $. This decomposition reduces the number of parameters by imposing a restriction on the rank of $W_i$ to be at most $k$ where $k \leq min(m,n)$. To further reduce the order of weight tensors by one, MLB replaces the unit tensor with another matrix $P_i \epsilon \mathbb{R}^{k \times c} $ to get,
\begin{equation}
    z_i = P_i^T(U_i^Tx \circ V_i^Ty)
\end{equation}
$k$ and $c$ are hyperparameters that decide the dimension of the joint embeddings and the dimension of the model output respectively. But MLB has problems too. It takes many iterations to converge and it is sensitive to hyperparameters.
\par
MFB\cite{mfb} modifies MLB by replacing $U \epsilon \mathbb{R}^{m \times k \times o} $ and $V \epsilon \mathbb{R}^{n \times k \times o} $ with $U' \epsilon \mathbb{R}^{m \times ko} $ and $V' \epsilon \mathbb{R}^{n \times ko} $ and then performing,
\begin{equation}
    z = SumPool(U'^Tx \circ V'^Ty, k)
\end{equation}
where the function $SumPool(X, k)$ means using a 1-D non-overlapping window with size $k$ to perform sum pooling over $X$. The authors show that MLB is a special case of MFB with $k = 1$. They also introduce MFH which can perform higher dimensional pooling by stacking multiple MFBs.
\par
MUTAN\cite{mutan} uses Tucker decomposition\cite{tuckerdecomp} to decompose $W$ into,
\begin{equation}
    W = \tau_c \times W_q \times W_v \times W_o
\end{equation}
Each of the four components have a specific role in the modeling. Matrices $W_q$ and $W_v$ project the question and the image vector into dimensions $t_q$ and $t_v$ respectively. These dimensions directly impact the complexity of modeling for each modality. Tensor $\tau_c$ controls the complexity of the interactions between projections produced by $W_q$ and $W_v$. The number of parameters can be controlled by imposing constraints on the rank of $\tau_c$. Finally, the matrix $W_o $ scores each pair of embedding for each class in the answer set. It can be shown that MCB and MLB are special cases of MUTAN.

\par
Finally, there is BLOCK\cite{block} which uses block-term decomposition. BLOCK aims to achieve a balance between MLB and MUTAN. MLB can be thought of as having the same number of blocks as the rank, each block being of size $(1, 1, 1)$. This means that projections of both modalities can be very high dimensional but interactions between them will be poor. In comparison, MUTAN uses only one block of size $(L, M, N)$ which means better interactions between projections but the projections are modeled less accurately. BLOCK takes a middle approach between these methods and performs better.

\begin{table*}[h!]
\centering
\begin{tabular}{ |M{3cm}|M{3cm}|M{3cm}| }
\hline
Model & VQA-v1 & VQA-v2 \\
\hline
MLB\cite{mlb} & 65.1 & 66.62 \\
\hline
MFB\cite{mfb} & 65.8 & - \\
\hline
MCB\cite{mcb} & 66.5 & 62.27 \\
\hline
MUTAN\cite{mutan} & 67.36 & 66.38 \\
\hline
BLOCK\cite{block} & - & 67.92 \\
\hline
\end{tabular}
\caption{Accuracy of different fusion models.}
\end{table*}

\begin{figure*}[ht]
\centering
\includegraphics[width=15cm,
  height=8cm,
  keepaspectratio]{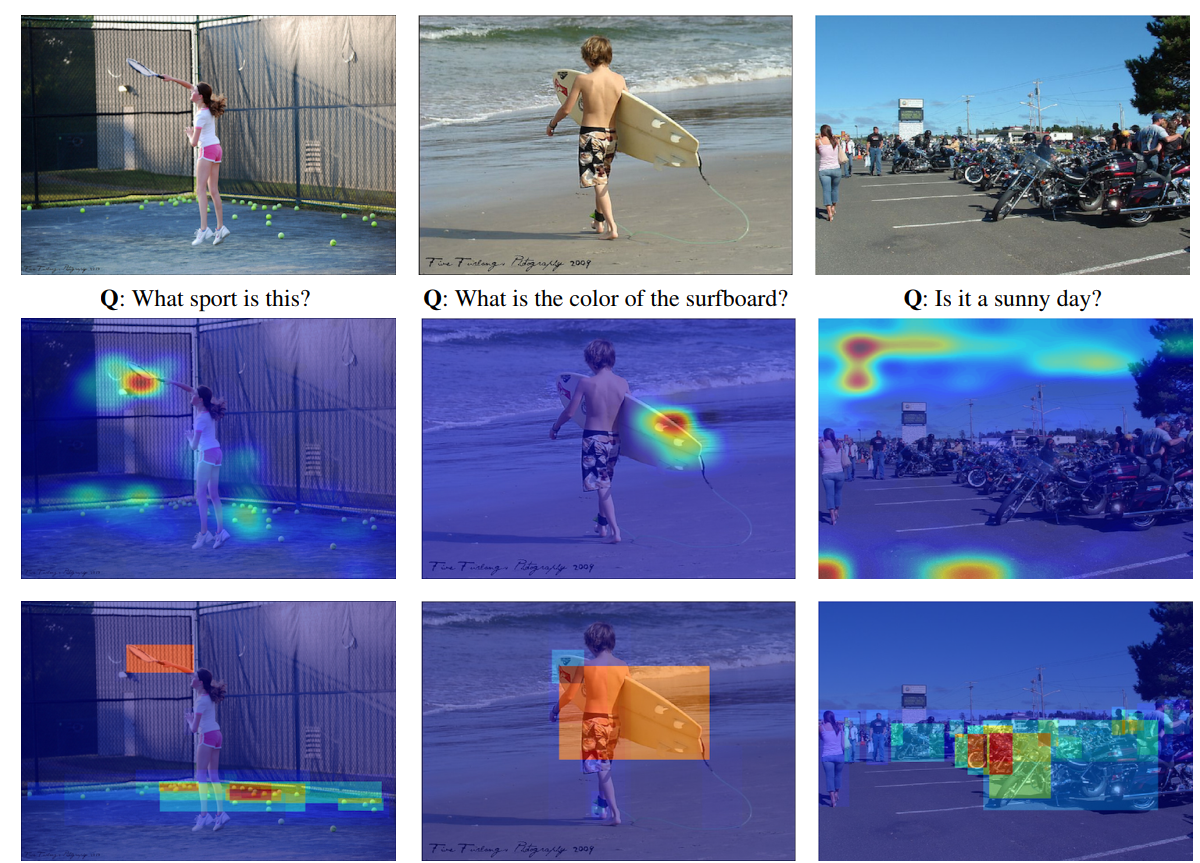}
\caption{First row presents the original images. Second row presents grid-based soft attention using CNNs. Third row presents object-based hard attention using object detectors. Figure from \cite{dual-mfa}}
\end{figure*}

\section{Attention}
Attention is the mechanism most widely used by VQA models to improve accuracy. When we are shown an image and asked a question about that image, we usually focus on the particular region of the image that helps us answer the question. VQA models try to replicate this through attention. This improves accuracy because a lot of irrelevant noise is filtered out and the model can focus better on the parts of the image that are actually relevant for answering the question. There are many different types of attention which we discuss below.

\subsection{Soft and hard attention}
The first aspect of note is soft and hard attention. Most models use the question to produce an attention map over the image. This map assigns different weights to different regions in the image according to their perceived relevance in answering the question. In soft attention, every region is assigned weights with more weight given to more relevant areas. In hard attention, \textit{only} regions which are deemed sufficiently relevant are selected for further consideration while all other regions are completely ignored. So, to compare, hard attention will completely ignore some areas of the image while soft attention will pay less attention to those areas. The selected regions in hard attention may go through a further soft attention stage.

\begin{table*}[t]
\centering
\begin{tabular}{ |M{3cm}|M{12cm}| } 
\hline
Soft, grid-based, top-down attention & SAN\cite{san}, SMem\cite{smem}, QAM\cite{qam}, MRN\cite{mrn}, RAU\cite{rau}, VQA-Machine\cite{vqa-machine}, HieCoAtt\cite{hiecoatt}, DAN\cite{dan}, MLAN\cite{mlan}, HOA\cite{hoa}, GVQA\cite{gvqa}, DCN\cite{dcn}\\
\hline
Hard, object-based, bottom-up Attention & Foc-reg\cite{foc-reg}, FDA\cite{fda}, BUTD\cite{butd}, AOA\cite{aoa}, CVA\cite{cva}, FEA\cite{fea}, MuRel\cite{murel}, DFAF\cite{dfaf}, MCAN\cite{mcan} \\
\hline
Both & Dual-MFA\cite{dual-mfa}, QTA\cite{qta}, DRAU\cite{drau} \\
\hline
\end{tabular}
\caption{Models with different attention schemes.}
\end{table*}

\begin{table*}[ht]
\centering
\begin{tabular}{ |M{3cm}|M{12cm}| } 
\hline
Single-step Attention & Foc-reg\cite{foc-reg}, QAM\cite{qam}, FDA\cite{fda}, VQA-Machine\cite{vqa-machine}, HieCoAtt\cite{hiecoatt}, MLAN\cite{mlan}, BUTD\cite{butd}, HOA\cite{hoa}, AOA\cite{aoa}, CVA\cite{cva}, QTA\cite{qta} \\
\hline
Multi-step Attention & SAN\cite{san}, SMem\cite{smem}, MRN\cite{mrn}, RAU\cite{rau}, DAN\cite{dan}, GVQA\cite{gvqa}, Dual-MFA\cite{dual-mfa}, DCN\cite{dcn}, FEA\cite{fea}, MuRel\cite{murel}, DFAF\cite{dfaf}, MCAN\cite{mcan}, DRAU\cite{drau} \\
\hline
\end{tabular}
\caption{Models using single and multi-step attention.}
\end{table*}

\subsection{Grid and object based attention}
Attention can also be divided into grid and object based approaches. In the grid based approach, the whole image is divided into a uniform grid where each cell is the same size. Almost all grid based approaches use soft attention. As the image is divided into regions of same size, many regions contain multiple objects. Using hard attention in this case may erase information about multiple objects which could result in significant information loss. In object based approaches, object detectors are used to divide the image into multiple objects. This falls under hard attention as image area that is not contained by the bounding box of any object is completely ignored. This is why some object based approaches also provide the whole image to the model to ensure access to all information.
\par
Some recent models\cite{dual-mfa}\cite{qta}\cite{raf} try to balance between grid and object based attention. They argue that some questions benefit more from object level attention and some benefit more from a grid based approach. These models extract image features from both grid-based CNN and object-based Faster R-CNN. The question is then used to assign weights to the features. In this way, the question itself determines which image feature should be used to answer the question.

\subsection{Bottom-up and top-down attention}
Another important concept is bottom-up and top-down attention. This is related to how human brains process vision. In the human visual system, attention can be focused volitionally by top-down signals determined by the current task, e.g., looking for something, and automatically by bottom-up signals associated with unexpected, novel, or salient stimuli\cite{humanbutd1}\cite{humanbutd2}. Similarly in the context of VQA, a top-down approach means considering the question first and then looking at the image while a bottom-up approach implies looking at the image first and then considering the question. BUTD\cite{butd} argues that both types of attention are useful. BUTD first applies bottom-up attention using Faster R-CNN to detect the most salient objects. This is equivalent to just looking at an image and noting all the prominent objects. BUTD then weighs these objects according to the question. This is top-down attention as visual information is being processed after considering the question.

\subsection{Single and multi-step attention}

\begin{figure}[ht]
\centering
\includegraphics[width=12cm,
  height=3cm,
  keepaspectratio]{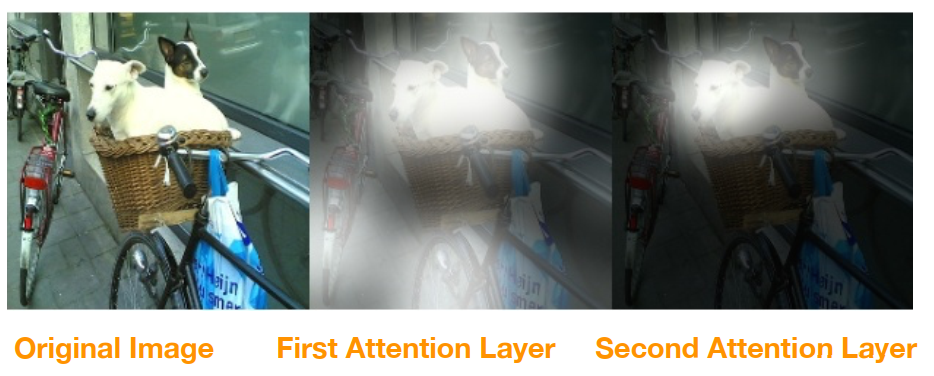}
\caption{Multi-step attention through multiple attention layers in SAN. Figure from \cite{san}}
\end{figure}

Attention can be single or multi-step. This is also referred to as “glimpse”. A model using multiple glimpses means multiple attention layers have been cascaded one after another. Some questions are simple and only require a single attention step to answer. For example, “What color is the dog?” only requires a single attention step to detect the dog. But many questions require multiple reasoning steps which may also require applying attention multiple times. For example, “What color is the dog standing beside the chair behind the counter?”. This question requires three attention steps. The first one to identify the \textit{counter}, the second to identify \textit{the chair behind the counter}, and the third to identify \textit{the dog beside the chair}. So we see that multi-step attention can lead to improved reasoning skills.

\begin{figure*}[ht]
\centering
\includegraphics[width=15cm,
  height=8cm,
  keepaspectratio]{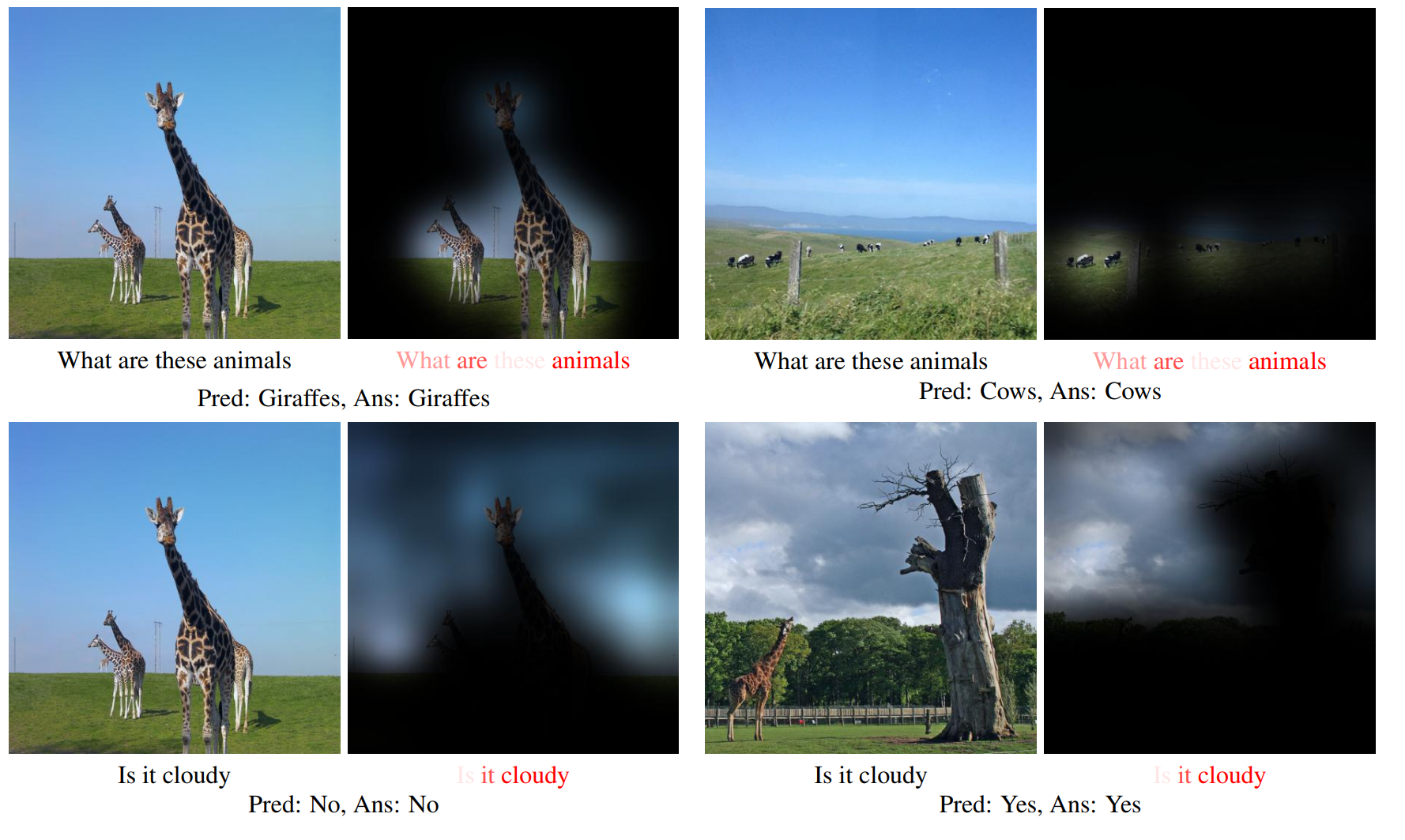}
\caption{Co-attention where both image and question guide attention on each-other. The rows present co-attention for two different images while the question is same. The columns present co-attention for the same image while the questions are different. Figure from \cite{dcn}}
\end{figure*}

\begin{table*}[t]
\centering
\begin{tabular}{ |M{3cm}|M{12cm}| } 
\hline
Co-attention(Image by question) &  Foc-reg\cite{foc-reg}, SAN\cite{san}, SMem\cite{smem}, QAM\cite{qam}, FDA\cite{fda}, MRN\cite{mrn}
, VQA-Machine\cite{vqa-machine}, HieCoAtt\cite{hiecoatt}, DAN\cite{dan}, MLAN\cite{mlan}, GVQA\cite{gvqa}, Dual-MFA\cite{dual-mfa}
, AOA\cite{aoa}, DCN\cite{dcn}, FEA\cite{fea}, MuRel\cite{murel}, DFAF\cite{dfaf}, DRAU\cite{drau}, MCAN\cite{mcan}\\
\hline
Co-attention(Question by image) &  HieCoAtt\cite{hiecoatt}, DAN\cite{dan}, BUTD\cite{butd}, DCN\cite{dcn}, DFAF\cite{dfaf}, DRAU\cite{drau}, MCAN\cite{mcan}\\
\hline
Self-attention(Image) &  DFAF\cite{dfaf}, MCAN\cite{mcan}\\
\hline
Self-attention(Question) &  DFAF\cite{dfaf}, MCAN\cite{mcan}\\
\hline
\end{tabular}
\caption{Models using co-attention and self attention.}
\end{table*}

\subsection{Co-attention and self-attention}
Until now we have only discussed image attention but what about attending the question?. HieCoAtt\cite{hiecoatt} was the first model to seriously explore question attention. They argued that just as image attention can filter out noise from the image, question attention can also reduce unwanted features. Just as the question can guide attention on the image, the image can also guide attention on the question. This is referred to as \textit{co-attention}.  Co-attention can take the image into context and apply weights to question words according to how relevant they are for answering \textit{that particular image}.

In co-attention, image and question attend each-other but it is also possible for each modality to apply attention to itself. This is aptly called self-attention. Self-attention can be applied to the question because not all words in a question are of equal importance. For example, consider the question “What color is the dog?”. Here the question can be shortened to “What color dog” without losing any significant meaning. “is” and “the” can be considered noise which do not contribute to the model's understanding of the question. Self-attention can also be applied to the image. Self-attention for image determines which region or object is important after taking into consideration all other regions and objects. For example, consider an image of a sheep-dog guarding a herd of sheep. Here the individual sheep are less important and the herd is more important as a whole. So self-attention can modify the weights such that the combined weight of all the sheep is equal to the weight of the dog which basically embodies the main relation in the image which is “A dog guarding a herd of sheep”.
\par

\begin{table*}[t]
\centering
\begin{tabular}{ |M{3cm}|M{3cm}|M{3cm}| } 
\hline
Model & VQA-v1 & VQA-v2 \\
\hline
SAN\cite{san} & 58.9 & - \\
\hline
FDA\cite{fda} & 59.5 & - \\
\hline
MRN\cite{mrn} & 61.8 & - \\
\hline
HieCoAtt\cite{hiecoatt} & 62.1 & - \\
\hline
RAU\cite{rau} & 63.2 & - \\
\hline
DAN\cite{dan} & 64.2 & - \\
\hline
MLAN\cite{mlan} & 64.8 & - \\
\hline
Dual-MFA\cite{dual-mfa} & 66.09 & - \\
\hline
BUTD\cite{butd} & - & 65.67 \\
\hline
DCN\cite{dcn} & 67.02 & 67.04 \\
\hline
DRAU\cite{drau} & 67.16 & 66.85 \\
\hline
MuRel\cite{murel} & - & 68.41 \\
\hline
DFAF\cite{dfaf} & - & 70.34 \\
\hline
MCAN\cite{mcan} & - & 70.90 \\
\hline
\end{tabular}
\caption{Accuracy of different attention models.}
\end{table*}

\section{External Knowledge}

Sometimes the information required to answer a question can not be found in the given image. For example, The question “What do the two animals in the image have in common?” for an image containing a Giraffe and a Lion can not be answered from the image alone. To correctly answer this question, i.e., they are both mammals, the model has to have an idea about the outside concept of “mammal”. As conventional VQA models have no way of gathering knowledge about unseen concepts, they will guess blindly from the answer space of known concepts. Most VQA models have limited knowledge of real-world concepts because the datasets they are trained on are restricted in their size and cannot reflect every combination of every concept. A simple way to solve this problem is to give the model the capability to query an External Knowledge Base or EKB.
\par
EKBs are structured representations of real-world knowledge. Construction, organization, and querying of these knowledge bases are problems that have been studied for years. This has led to the development of large-scale KBs constructed by human annotation, e.g., DBpedia\cite{dbpedia}, Freebase\cite{freebase}, Wikidata\cite{wikidata}) and KBs constructed by automatic extraction from unstructured or semi-structured data, e.g., YAGO\cite{yago2}\cite{yago3}, OpenIE\cite{openie1}\cite{openie2}\cite{openie3}, NELL\cite{nell}, NEIL\cite{neil}, WebChild\cite{webchild}, ConceptNet\cite{conceptnet}. In structured KBs, knowledge is typically represented by a large number of triples of the form (arg1, rel, arg2). “arg1” and “arg2” denote two Concepts in the KB, each describing a concrete or abstract entity with specific characteristics. “rel” represents a specific Relationship between them. A collection of such triples forms a large interlinked graph. Information can be retrieved by using an SQL-like query language. Queries usually return one or multiple sub-graphs satisfying the specified conditions.
\par

\begin{figure}[ht]
\centering
\includegraphics[width=8cm,
  height=8cm,
  keepaspectratio]{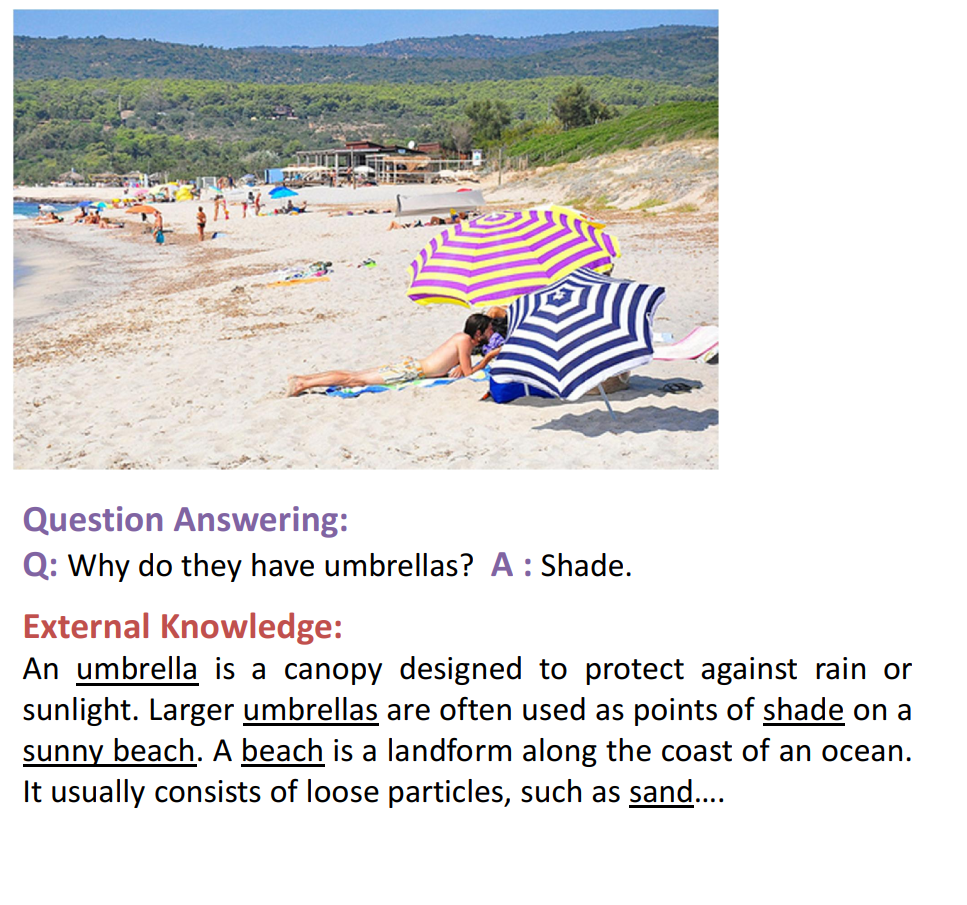}
\caption{Example of an image-question pair that needs external knowledge to answer. Figure adapted from \cite{ama}}
\end{figure}

\begin{figure}[ht]
\centering
\includegraphics[width=10cm,
  height=6cm,
  keepaspectratio]{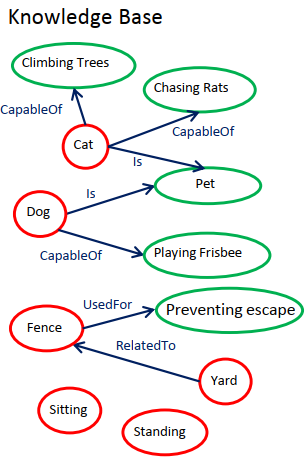}
\caption{An example structure of an EKB. Figure from \cite{fvqa}}
\end{figure}

EKB models usually try to map image concepts and question concepts to their equivalent KB concepts as extra information about those concepts is likely to be helpful in the answering process. Some models\cite{ahab},\cite{fvqa} do this by building an image-specific knowledge graph. Visual concepts such as objects, scenes, attributes, actions, etc., are detected in the image and used to build a scene graph. A knowledge sub-graph is produced by querying the knowledge base with the detected image concepts. The two graphs are combined by linking the visual concepts with their corresponding KB concepts. This combined graph is the image-specific knowledge graph that integrates additional information about various image content. Along with visual concepts, KDMN\cite{kdmn} also uses keywords from the question to query the knowledge base. Thus, its knowledge graph contains question concepts along with visual concepts. ACK\cite{ack} uses the image to predict top-5 attributes and uses them to query DBpedia which returns five knowledge paragraphs. STTF\cite{sttf} and OOTB\cite{ootb} extract visual concepts from the image and use the question to filter relevant facts from a facts database. STTF then selects a top fact by a ranking mechanism and uses this to answer the question whereas OOTB\cite{ootb} constructs a graph and uses graph convolution to produce the final answer.
\par
Although there has been significant work in integrating external knowledge in VQA, EKB models still suffer from many issues. Some of these issues stem from the difficulty of interacting with most KBs. KBs cannot be queried with natural language questions and need specific query formats. There are only a limited number of ways in which a KB can be queried and any question we want to ask the KB must be reducible to one of the available query templates. Parsing the image and the question to produce appropriate queries is still an error-prone process that many EKB models struggle with. Another issue is with integrating knowledge graphs. They are symbolic in nature which makes it difficult to train an EKB model in an end-to-end differentiable manner. Some models work around this by converting the knowledge graph into a continuous vector representation. For example, KDMN\cite{kdmn} treats each knowledge triple as a three-word SVO(subject-verb-object) phrase. It embeds each triple into a common feature space by feeding its word embeddings through an RNN architecture. In this case, the external knowledge is embedded in the same semantic space as the question and the answer. Thus, the model can be trained in an end-to-end manner. 

\section{Compositional Reasoning}
Most VQA models do not possess significant reasoning skills. They show good performance on general VQA datasets which lack complex questions. They fail on datasets like CLEVR which was designed specifically to test a model's high level reasoning skill. To correctly reason about complex questions, a model needs to be compositional in nature. By composition, we refer to the ability to break a question down into individual reasoning steps which when done sequentially produces the correct answer. An example of such a question would be “How many large cubes are behind the purple cylinder?”. Answering this question requires detecting the purple cylinder, then filtering the objects behind it, then filtering again to find objects that are both large and cube-shaped, and finally counting them. In contrast, most VQA models are trained to answer simple questions which usually require a single visual verification step. We divide compositional models into two camps, 1) PGE(Program generator and executor) models and 2) Blackbox models.

\subsection{PGE models}
PGE models consist of two parts, program generation and program execution. In the first part, the input question is parsed to produce a program. This program is executed in the second part to obtain the answer. Most complex questions can be mapped to a dependency tree where each node represents a single reasoning step. The correct answer can be found by traversing the tree and processing each node appropriately. After a question has been mapped to a valid tree, the next step is to transform this tree into an executable program by using neural modules. PGE models have a toolbox of neural modules where each module can perform a specific reasoning task such as a “find” module that can locate an object/concept in the image. A neural module is plugged into each node of the tree according to the task that node represents. It is important to note here that these modules are not hard-coded but their parameters are trained just like a neural network. In the next step, the complete network is dynamically assembled by laying out the modules in the tree. The answer is predicted by feeding the inputs to this network. Due to their modular nature, PGE models offer greater transparency and interpretability. As the inference process follows a chain of discrete steps, reasoning failures can be pinpointed to a particular node which is not possible in a traditional end-to-end neural network. This can also be leveraged to produce explanations in order to validate sound reasoning.

\begin{table*}[t]
\centering
\begin{tabular}{ |M{3cm}|M{3cm}|M{3cm}|M{3cm}| } 
\hline
Model & VQA-v1 & VQA-v2 & CLEVR \\
\hline
NMN\cite{nmn} & 58.7 & - & - \\
\hline
D-NMN\cite{d-nmn} & 59.4 & - & - \\
\hline
N2NMN\cite{n2nmn} & - & 63.3 & 83.7 \\
\hline
PTGRN\cite{ptgrn} & - & - & 95.47 \\
\hline
Stack-NMN\cite{stack-nmn} & 66.0 & 64.0 & 96.5 \\
\hline
IEP\cite{iep} & - & - & 96.9 \\
\hline
DDRProg\cite{ddrprog} & - & - & 98.3 \\
\hline
NS-CL\cite{ns-cl} & - & - & 98.9 \\
\hline
TbD-NET\cite{tbd-net} & - & - & 99.1 \\
\hline
UnCoRd\cite{uncord} & - & - & 99.74 \\
\hline
NS-VQA\cite{ns-vqa} & - & - & 99.8 \\
\hline
XNM\cite{xnm} & - & 67.5 & 100 \\
\hline
\end{tabular}
\caption{Accuracy of different PGE models.}
\end{table*}

\begin{figure}[ht]
\centering
\includegraphics[width=8cm,
  height=6cm,
  keepaspectratio]{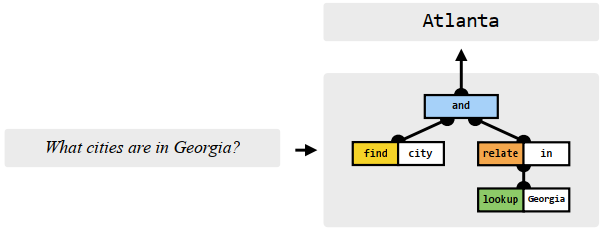}
\caption{An example of a tree parsed from a question. Figure adapted from \cite{d-nmn}}
\end{figure}

\par
PGE models have to parse the input question into a dependency tree which can be used to produce a module layout. Earlier models such as NMN\cite{nmn} and D-NMN\cite{d-nmn} used the Stanford dependency parser to produce the dependency tree. But due to the parser being error-prone, the selected layout was often wrong. Later models\cite{n2nmn}\cite{stack-nmn}\cite{xnm} treated layout mapping as a sequence-to-sequence learning problem and replaced off-the-shelf parsers with encoder-decoder models. At first, models like N2NMN\cite{n2nmn} selected layouts discreetly which was a non-differentiable process. This meant that the program generation part of the model had to be optimized using reinforcement learning while the execution part was trained using gradients. This multi-stage joint training was difficult as the program generator needed to produce the right program without understanding what the individual modules meant, and the execution engine had to produce the right answer from a program that might not accurately represent the question. Later models like Stack-NMN\cite{stack-nmn} and XNM\cite{xnm} converted layout selection into a soft continuous process so that the model could be optimized in a fully differentiable manner using gradient descent. DDRprog\cite{ddrprog} solved the problem of differentiability by interleaving module prediction and generation, thereby producing the layout “on the fly”. PTGRN\cite{ptgrn} solves it by eliminating the need for module prediction. They do this by using generic PTGRN modules instead of specialized neural modules. They directly use the dependency tree produced by the Stanford parser as the program layout by replacing each node with a PTGRN node module and each edge with a PTGRN edge module. 

\par
Earlier PGE models could not generalize well to mainstream VQA datasets as they required program supervision from layout annotations. More recent models like XNM do not require expert layouts and have shown competitive performance on general VQA datasets on par with models that do not use compositional reasoning. This is significant considering that non-compositional models have been shown to utilize various dataset biases while PGE models are less prone to bias due to their modular architecture.

\begin{table*}[t]
\centering
\begin{tabular}{ |M{3cm}|M{3cm}| } 
\hline
Model & CLEVR \\
\hline
FiLM\cite{film} & 97.7 \\
\hline
CMM\cite{cmm} & 98.6 \\
\hline
MAC\cite{mac} & 98.9 \\
\hline
OCCAM\cite{occam} & 99.4 \\
\hline
\end{tabular}
\caption{Accuracy of different Blackbox models.}
\end{table*}

\subsection{Blackbox models}
A major setback for many PGE models is that they rely on program annotations, brittle handcrafted parsers, or expert demonstrations and require relatively complex multi-stage reinforcement learning training schemes. These models are also limited to using a few handcrafted neural modules each with a highly rigid structure. This makes many PGE models inflexible compared to conventional neural network-based VQA models. Blackbox models try to achieve a balance between both worlds. Like PGE models, blackbox models utilize structural priors to a certain extent while retaining the fluidity and dynamic nature of neural network-based models. Blackbox models achieve this by using a single highly flexible blackbox module instead of multiple operation-specific neural modules. Multiple blackbox cells are cascaded one after another to form a long chain. Because of self-attention and residual connections between the cells, this linear network is capable of representing arbitrarily complex acyclic reasoning graphs in a soft manner. Because of the use of a single reusable module and the physically sequential nature of the network, blackbox models manage to circumvent two major bottlenecks of PGE models, namely, module selection and layout selection. This allows blackbox models to be trained in an end-to-end manner using simple backpropagation just like conventional VQA models instead of the complex multi-stage training of PGE models. Though multiple instances of the same cell are used in multiple layers, they only share the general structure. Through training, different cells in different layers learn to perform different reasoning operations. This allows blackbox models to perform long and complex chains of reasoning. Blackbox models retain fluidity and flexibility while maintaining compositional reasoning ability albeit at the cost of some transparency and interpretability.

\par
MAC\cite{mac} consists of a sequence of MAC cells. MAC follows a controller-memory scheme where strict separation is maintained between the representation space of the question and the image. This is in contrast to the conventional approach of fusing the question and the image representation in the same semantic space. Each MAC cell consists of three units: a) Control; b) Read; and c) Write. They operate on a control state and a memory state. The control states are determined by the question and the control states in different cells are expected to encode a series of reasoning operations. The read unit extracts the required visual information from the image guided by the control and memory state. After the control unit performs its reasoning operation, the write unit integrates the new information into the memory state. This architecture allows MAC to divide its operation into discrete reasoning steps where the question and the image interact in a controlled way. This makes it easier to interpret the inner workings of a particular MAC cell resulting in better transparency. Similar to MAC, FiLM\cite{film} uses the question to guide its computation over the image. In contrast to MAC, where the question does not influence the image feature extraction process, FiLM uses the question to guide how its visual pipeline extracts features from the image. FiLM consists of several convolution layers like a typical CNN. But in FiLM, each convolution layer is paired with a FiLM layer which controls the feature map computation in that layer. The parameters of each FiLM layer are determined by the question. The intuition is that each FiLM layer will encode the reasoning operation necessary at that stage of the pipeline. CMM\cite{cmm} modifies FiLM so that both modalities are influenced by each other. It argues that, since information comes from multiple modalities, it is not intuitive to assume that one modality(language) is the “program generator”, and the other(image) is the “executor”. One way to avoid making this assumption is to perform multiple steps of reasoning with each modality being generator and executor alternatively in each step.

\section{Explanation}

A recent trend in VQA models has been towards explainability. Many have argued that VQA models should be able to explain their answers, i.e., the precise reasoning steps the model followed to deduce the answer. Traditional neural networks act like blackboxes whose inner workings remain opaque to outside observers. This makes it difficult to verify if the network is reasoning correctly. The model may be using dataset biases or merely guessing the answer. There is no way to be sure that the model is producing the right answers for the right reasons. A good AI model should be able to explain its actions and expose the reasoning process it followed.  Forcing models to generate explanations ensures trust in VQA systems and exposes models which merely use biases and shortcuts. Explanations force models to reveal their internal process and it can be verified that the correct answer is being predicted logically. In case of the model predicting a wrong answer, the precise step where failure occurred can be pinpointed. Maybe the model failed to understand image content correctly or maybe the model had trouble understanding the question. These are understanding failures. It could also be that the model understood the image and the question properly but failed to follow the correct reasoning steps. This is a reasoning failure. Explanations help to gain a better understanding of how the model is actually operating. So if the model predicts the correct answer, we can make sure it was because it followed proper logical reasoning. If the prediction is wrong, we can narrow down the failed step. Generating explanations acts as a type of regularization method. It results in better answer prediction as the model focuses on proper reasoning instead of taking shortcuts.

\begin{figure}[ht]
\centering
\includegraphics[width=8cm,
  height=6cm,
  keepaspectratio]{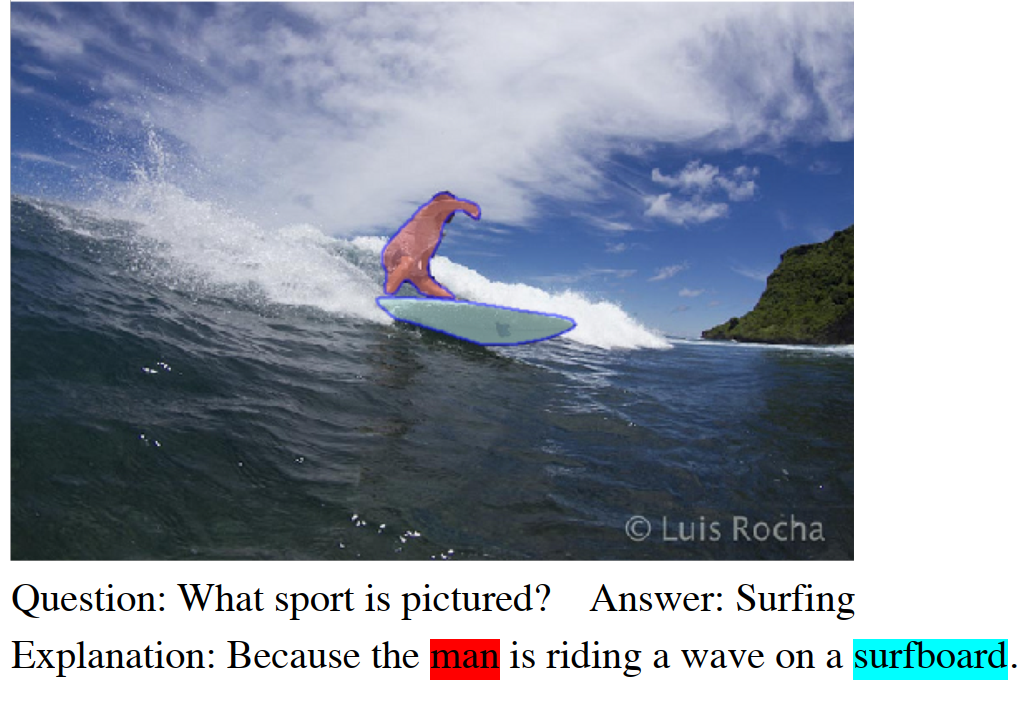}
\caption{Example of both visual explanation and textual explanation. Figure from \cite{faithfulexplanation}}
\end{figure}

\par
Explanations can be visual or textual. Visual explanations are attention heatmaps indicating which image regions the model focused on for answer prediction. Textual explanations are natural language explanations usually comprising of multiple sentences which describe key information the model used in its reasoning process. Earlier explanation models only produced visual explanations as attention heatmaps are easy to build using gradient based methods such as Grad-CAM\cite{gradcam}. Visual explanations are intuitive and reveal the areas of the image the model concentrated on. If the model deduced the right answer but did not look at the right area, the model's reasoning is suspect. If the model failed to answer correctly, we can analyse the attention map to figure out why the model failed. But visual explanations alone may not be enough to fully understand how the model is operating. Many times the model attends to the correct region but still fails to use the visual information correctly in its inference process. Textual explanations can provide better insights and reveal whether the attended visual information was used correctly. Natural language explanations are also easier to understand for humans. Textual explanations can explain the reasoning process step by step which is clearer than a single attention map. The best approach seems to be producing both types of explanations. Models generating such multimodal explanations argue that visual and textual explanations complement each-other. Both fine-tune each other and improve each other's quality. Generating natural language explanations results in better attention maps making them less noisy, more relevant and focused. On the other hand, attending to important image regions motivates the textual explanations to be more relevant, more succinct and more related to image content.

\par
\cite{interpretable} uses attention supervision to train the model to look at relevant image regions and produce an attention map of the image as the visual explanation. It devises a way to automatically collect this attention supervision from available region descriptions and object annotations. \cite{tellandanswer} divides it's answering process into two parts. In the “explaining” part, the image is used to detect attributes and produce a caption. These are used in the “reasoning” part in place of the image to infer the correct answer. As the model bases its reasoning on the produced caption and attributes, they can be thought of as textual explanations explaining what information the model deduced its answer from. \cite{vqa-e} uses human-generated explanations to supervise its textual explanation generation. It argues that forcing the model to produce natural language explanations can help it predict more accurate answers. \cite{explanationscene} uses the attention heatmap to separate relevant parts of the image scene graph. It then uses this structured information about the relevant scene entities to produce suitable natural language explanations. \cite{pj-x} produces multimodal explanations which point to the visual evidence for a decision and also provides textual justifications. \cite{faithfulexplanation} argues for more faithful multimodal explanations where both visual and textual explanations ground and regulate each-other. Prominent objects in the attention map should be mentioned in the explanation text and the textual explanation should ground any object it mentions in the image. In this way, both explanations calibrate each-other resulting in less noise and more relevant explanations. 

\section{Graph models}

Graph models use graphs to process information from the image and the question. Graph models argue that conventional VQA models cannot do complex reasoning with their monolithic vector representations. Graph models pin the ability of a model to do higher-level reasoning on its capacity to perform relational reasoning. Almost any complex question about an image requires reasoning about its objects and relationships. For example, a ‘boy’ and a ‘sandwich’ with the relation ‘eating’ encodes an important fact about an image. Another fact like this could be (‘boy’, ‘next to’, ‘dog’). Now if a VQA model is asked the question ‘What is the boy next to the dog eating?’, it will be much more successful in answering the question if it understands the various relationships that exist among the multiple objects. As we will see, graph models use graphs to model these relationships and use them to answer questions.
\par
There can be two types of graphs in a graph model, an intra-modality graph and an inter-modality graph. Intra-modality graphs are graphs belonging to a single modality, either the image or the question. In image graphs, the nodes represent the objects detected in the image and the edges represent the relationships between them. In question graphs, the nodes represent the words and the edges could represent their syntactic dependencies or some other relationship. Having two separate graphs means that the image and the question do not interact with each other but rather their elements interact within themselves. In this way, intra-modality graphs can model self-attention or simply produce better image and question features. On the other hand, inter-modality graphs model interactions between the two modalities. These interactions could be direct(fusion) or indirect(attention).

\begin{table*}[t]
\centering
\begin{tabular}{ |M{3cm}|M{3cm}|M{3cm}| } 
\hline
Model & VQA-v2 & GQA \\
\hline
CGS\cite{cgs} & 66.18 & - \\
\hline
MuRel\cite{murel} & 68.41 & - \\
\hline
LCGN\cite{lcgn} & - & 56.1 \\
\hline
ReGAT\cite{regat} & 70.58 & - \\
\hline
GRN\cite{grn} & 71.12 & 57.04 \\
\hline
TRRNet\cite{trrnet} & 71.2 & 60.74 \\
\hline
BGN\cite{bgn} & 72.41 & - \\
\hline
NSM\cite{nsm} & - & 63.17 \\
\hline
\end{tabular}
\caption{Accuracy of different Graph models.}
\end{table*}

\par
Graph-VQA\cite{graph-vqa} builds two graphs, a question graph and a scene graph. In the question graph, nodes are word features and edges are syntactic dependency types as predicted by the Stanford Dependency Parser. In the scene graph, nodes are object features and edges are relative spatial positions. LCGN\cite{lcgn} and TRRNet\cite{trrnet} also use object features as nodes in their graphs. While LCGN uses all detected objects, TRRNet selects top k objects using the question. NSM\cite{nsm} generates a scene graph where each object node is accompanied by a bounding box, a mask, dense visual features, and a collection of discrete probability distributions for each of the object’s semantic properties, such as its color, material, shape, etc., defined over a fixed concept vocabulary. Some graph models do not represent objects as nodes. Rather they fuse the object features with the question\cite{cgs}\cite{murel}\cite{regat} or the words\cite{bgn} and treat the joint embeddings as individual nodes.
\par
A Graph model usually processes its graphs in multiple iterations. In each iteration, each node gathers context and information from its neighbors and updates itself. The edges could also update alongside the nodes\cite{lcgn} but most models only implement node updates. Each node has its own neighborhood where neighborhood could mean all other nodes or a subset of them based on some criterion. For example, the criterion could be that a node’s neighbors are the top k nodes according to edge weights. The neighborhood selection criterion is important due to computational concerns. A fully connected graph where each node considers all other nodes its neighbors is computation-heavy. On the other hand, a more choosy selection criterion might leave out relevant nodes and miss important relations and interactions. In most graph models like Graph-VQA\cite{graph-vqa} and NSM\cite{nsm}, each node considers its adjacent nodes to be neighbors. In CGS\cite{cgs}, a node’s neighbors correspond to the K most similar nodes. In other graph models like MuReL\cite{murel} and LCGN\cite{lcgn}, the graph is fully connected and each node considers all other nodes its neighbors.
\par
Graph processing can be done in three ways – graph convolution, graph attention, or graph traversal. We discuss these three methods through three models that each make use of one of them. CGS\cite{cgs} uses graph convolution. Graph convolution can be thought to summarize information from a particular neighborhood of nodes. Multiple layers of convolution can produce higher and higher levels of representations of the graph. CGS implements a patch operator by using a set of K Gaussian kernels of learnable means and covariances. For a given vertex, the output of the patch operator can be thought of as a weighted sum of the neighboring features. ReGAT\cite{regat} uses graph attention. In the final step of processing, it uses self-attention to produce attention weights that are used to compute the final features. ReGAT argues that compared to graph convolutional networks which treat all neighbors the same, graph attention assigns different weights to different neighbors and this can reflect which relations are more relevant to a specific question. Finally, we have graph traversal which is used by NSM\cite{nsm}. NSM converts the question into a set of instructions and treats the graph as a state machine where the nodes correspond to states and the edges correspond to transitions. Then it performs iterative computations where the question-generated instructions are treated as input and traversing the graph simulates traversing the state machine’s states effectively performing sequential reasoning. At each instruction step, NSM traverses the graph by using the current nodes and edges to determine the next set of nodes. Node features are not updated in graph traversal. Traversal just determines the final set of nodes the answer will be based on. After processing, some models\cite{cgs}\cite{murel}\cite{graph-vqa} summarize the node features in some way to represent the entire final graph. The summarization can be done by max-pooling or averaging the top k nodes or by predicting attention weights on the final nodes or some other method. This final representation is then used with the question to predict the answer.

\begin{table*}[t]
\centering
\begin{tabular}{ |M{3cm}|M{7cm}| } 
\hline
Models &  Pre-training Tasks\\
\hline
LXMERT\cite{lxmert} &  MLM, MRFR, MRC, ITM, IQA\\
\hline
ViLBERT\cite{vilbert} &  MLM, MRC, ITM\\
\hline
VisualBERT\cite{visualbert} &  MLM, ITM\\
\hline
VL-BERT\cite{vl-bert} &  MLM, MRC\\
\hline
UNITER\cite{uniter} &  MLM, MRFR, MRC, ITM, WRA\\
\hline
VLP\cite{vlp} &  MLM, Uni-MLM\\
\hline
Oscar\cite{oscar} &  MLM, Contrastive Loss\\
\hline
Pixel-BERT\cite{pixel-bert} &  MLM, ITM\\
\hline
ERNIE-ViL\cite{ernie-vil} &  MLM, MRC, ITM, SGP\\
\hline
\end{tabular}
\caption{Pre-training tasks used by different transformer models.}
\end{table*}

\begin{table*}[t]
\centering
\begin{tabular}{ |M{3cm}|M{3cm}| } 
\hline
Models &  VQA-v2\\
\hline
VLP\cite{vlp} &  70.7\\
\hline
ViLBERT\cite{vilbert} &  70.92\\
\hline
VisualBERT\cite{visualbert} &  71.00\\
\hline
VL-BERT\cite{vl-bert} & 72.22\\
\hline
LXMERT\cite{lxmert} &  72.54\\
\hline
Oscar\cite{oscar} &  73.82\\
\hline
UNITER\cite{uniter} &  74.02\\
\hline
Pixel-BERT\cite{pixel-bert} &  74.55\\
\hline
ERNIE-ViL\cite{ernie-vil} &  74.93\\
\hline
\end{tabular}
\caption{Accuracy of different transformer models.}
\end{table*}

\section{Transformer Models}
Many recent VQA models have utilized BERT-like transformer models to extract cross-modal joint embeddings. \cite{transformer} introduced transformers which use iterative self-attention to model the dependency of all input elements. BERT\cite{bert} is a transformer based model that is pre-trained on several proxy tasks to learn better linguistic representations. Inspired by BERT's success on a wide range of NLP benchmarks, recent efforts have tried to leverage task-agnostic pre-training to tackle Vision-and-Language(V+L) challenges. Most V+L tasks rely on joint multimodel embeddings to bridge the semantic gap between the visual and textual elements in image and text. Tranformer models try to learn this joint representation through pre-training similar to BERT. Instead of using separate vision and language pre-training, unified multi-modal pre-training allows the model to implicitly discover useful alignments between both sets of inputs and build up a useful joint representation. BERT-like models define some proxy tasks to enable this pre-training. The tasks involve masking one part of the data and training the model to predict the ‘missing’ part from the context provided by the remaining data. Through these proxy tasks, the model gains a deeper level of semantic understanding.
\par

Transformer models fall into two camps, two-stream models and single-stream models. In the two-stream architecture, two seperate transformers are applied to image and text and their outputs are fused by a third Transformer in a later stage. In contrast, single-stream models use a single transformer for joint intra-modal and inter-modal interaction. ViLBERT\cite{vilbert}, LXMERT\cite{lxmert}, and ERNIE-ViL\cite{ernie-vil} use the two-stream architecture while VisualBERT\cite{visualbert}, VL-BERT\cite{vl-bert}, UNITER\cite{uniter}, VLP\cite{vlp}, Oscar\cite{oscar}, and Pixel-BERT\cite{pixel-bert} use the single-stream method. Two-stream models argue that using separate transformers to perform isolated self-attention brings more specific representations while single-stream models argue that self-attention benefits more from the context provided by the other modality resulting in better cross-modal alignment. 
\par
Designing proxy tasks is the most important part of pre-training. The proxy tasks must force the model to use information from one modality to make valid inferences about the other. In the process, the model is expected to mine useful cross-modal representations that encode general semantic information. We describe some of these tasks. In MLM(Masked Language Modelling), some textual input tokens are randomly masked and the model has to figure out these tokens by using information from the remaining textual and visual tokens. Only using textual information often results in ambiguity which can only be solved by utilising image information. This forces the model to learn useful alignments between the modalities. By definition, MLM provides bi-directional context where all textual tokens that were not selected for masking are visible. VLP\cite{vlp} introduces an additional task in the form of unidirectional MLM where only the tokens on the left of the ‘to-be-predicted’ token are visible to the model. In MRFR(Masked Region Feature Regression), the model has to regress region features of randomly masked image regions. This task is harder to train on because regressing on a pixel level is difficult and visual ambiguity is more abundant than language ambiguity. For example, there are endless variations of valid pixel arrangements for a region labeled ‘dog’. MRC(Masked Region Classification) is a more tenable task where the labels of randomly selected image regions are masked and the model has to predict the missing object labels. UNITER\cite{uniter} introduces the WRA(Word-Region Alignment) task where Optimal transport is used to explicitly encourage fine-grained alignment between words and image regions. Optimal transport tries to minimize the cost of transporting the embeddings from image regions to words in a sentence and vice versa. In addition to text and image, Oscar\cite{oscar} provides a third input in the form of object tags. Oscar argues that object tags can act as helpful anchors to ease the learning of cross-modal alignments. Oscar defines a contrastive loss task where the model has to match objects tags to valid images. ERNIE-ViL\cite{ernie-vil} introduces scene graph prediction where the model is provided with three tasks - object prediction, attribute prediction, and relationship prediction. ERNIE-ViL contends that MLM pre-training alone cannot help the model determine whether a word represents an object or an attribute or a relation and that scene-graphs have to be explicitly included in the training objective for the model to gain better understanding. Pixel-BERT\cite{pixel-bert} argues that using region features from pre-trained object detectors causes an information gap because useful information like shape, spatial position, scene information, etc., are lost. Instead of region features, Pixel-BERT uses raw pixels as the visual input. To make this manageable, a random sampling method is used. This makes training computationally feasible and encourages the model to learn semantic knowledge from incomplete visual input which enhances robustness.
\par

\section{Miscellaneous}

\subsection{Understanding text in image}
Many questions about images relate to text present in the image. For example, “What does the sign say?”, “What time does the clock show?”, etc. Questions like this require new answers unseen during training. As it is not possible to include all possible answers in the test set, the model needs to have the ability to generate novel answers. Most of the existing VQA models are inept at reasoning about scene text. But some recent models have tried to tackle this problem. LoRRA\cite{textvqa} is a VQA model which has an additional OCR module that allows it to answer questions about scene text. The OCR module extracts all texts in the image as tokens and LoRRA determines which of these tokens is needed in the answering process. It can also output one of the tokens as the answer. ST-VQA\cite{stvqa}  is another model that tries to incorporate scene text understanding. ST-VQA argues that PHOC\cite{phoc} should be used for embedding the extracted text instead of traditional word embedding models. Unlike other word embedding models, PHOC puts more emphasis on the morphology of words rather than the semantics. As a result, it is more suitable for representing out-of-vocabulary words such as, license plate numbers.

\subsection{Counting}
Among the various sub-problems of VQA, counting is often singled out because of it's difficulty. Models that use CNNs for image features are prone to failing on counting questions. This is because the convolution and pooling layers of CNN aggregate spatial information of various objects together. The resulting features can identify the presence of a certain object but cannot count how many instances of that object are there. So, more recent models have switched to using object detectors. Object detectors can detect all instances of a particular object. The only issue here is that the predicted bounding boxes sometimes overlap which can lead to duplicate counting. Another issue is the information loss that occurs in the attention layer. Almost all models use normalized soft attention which sums up the attention weights to equal $1.0$. So if an image contains $1$ cat, that cat will receive an attention weight of $1.0$ but if there are $2$ cats both will receive weights of $0.5$ which effectively means that the two cats are being averaged back into a single cat. So, instead of all instances of the object getting the highest attention, attention gets divided and diluted. Skipping normalization is not an option as \cite{tipsandtricks} shows it degrades performance on non-counting questions while the increase in accuracy for counting questions is not that significant. 
\par
IRLC\cite{irlc} defines counting as a sequential object selection problem. IRLC uses reinforcement learning to train a scoring function based on the question. The Object with the highest score gets selected and this selection prompts new scores for the remaining objects. But the use of reinforcement learning hinders assimilating this method in existing VQA models. Counter\cite{counter} is a model that uses object detection to construct a graph of all detected objects based on how they overlap. Edges in the graph are removed based on several heuristics to ensure that each object is only counted once. RCN\cite{tallyqa} is a model made specifically for the VQA counting task. It uses Faster R-CNN to detect foreground(fg) and background(bg) objects. It then performs relational reasoning with all fg-fg and fg-bg object pairs. RCN argues that using relational reasoning instead of directly working with the bounding boxes allows it to overcome the overlapping issue.
 
 \subsection{Bias Reduction}
Dataset bias has been a persistent issue in VQA from the start. Many papers\cite{analysis1}\cite{analysis2}\cite{vqa-v2}\cite{clevr}\cite{yinyang}\cite{gvqa} have produced findings that suggest that most of the existing models in VQA increase their performance by utilising spurious correlations such as, statistical biases and language priors instead of properly reasoning about the image and question. Question bias in particular seems to plague most models making them indifferent to image information. Existing VQA datasets contain extremely skewed question and answer distributions. So it's hard to distinguish whether a model is really doing well or if it's just using these biases. To diagnose bias-reliance in models, VQA-cp\cite{vqacp} was introduced. It re-organizes the VQA dataset(both v1 and v2) to make the training and test splits contain different distributions. So only models with actual generalization ability do well on the test set while biased models fail. There have been many recent efforts at bias reduction in VQA. Approaches fall roughly under three categories: 1) Utilizing a bias-only model with the main VQA model; 2) Forcing the model to take the image into consideration; and 3) Augmenting new counter-factual samples. The first two approaches try to remove bias from the model side while the third tries to remove bias from the dataset side.
\par
A bias-only model can be fitted with the main VQA model to isolate training signals that originate from bias. \cite{adv-reg}, \cite{ensemble}, and \cite{rubi} follow this approach. \cite{adv-reg} uses adversarial training where they set the main VQA model and the question-only adversary against each other. The main VQA network tries to modify its question embedding so that it becomes less useful to the question-only model. This results in the question encoder capturing less bias in it's encoding. \cite{ensemble} follows the straight-forward route of using a simple ensemble of the main and the bias-only model. They argue that the ensembled model learns to use the unbiased model for questions that the biased model fails to answer. \cite{rubi} uses the biased model to dynamically adjust it's loss such that the network gives less importance to the most biased examples and more importance to the less biased.
\par
Another way to reduce question bias is to make sure the model takes the image into consideration. HINT\cite{hint} argues that the model may focus on the right region in the attention stage but still disregard the visual information from it's reasoning process. HINT works on a gradient level to make sure the answering process is affected by the visual branch of the model. HINT proposes a ranking loss which tries to align it's visual attention with human provided attention annotations. In contrast, SCR\cite{scr} introduces a self-critical training objective which penalizes the model if correct answers are more sensitive to non-important regions than to important regions, and if incorrect answers are more sensitive to important regions than correct answers.
\par
Some approaches have tried debiasing from the data side by augmenting existing data with counter-factual samples. These samples help the model to locate the underlying causal relations that lead to the correct answer. A counter-factual sample is produced by slightly modifying the original sample which causes a subtle semantic shift resulting in a different answer. Take for example an image of a dog. If the dog in the image is erased or masked, the answer to the question “Is there a dog in the image?” changes from “yes” to “no”. Usually counter-factual samples differ from the original sample in a slight way while still remaining very similar. This forces the model to gain a more nuanced understanding in order to differentiate between the two samples. For VQA, counter-factual samples are produced by modifying the image or the question or both. Images can be modified by masking one or more regions. Masking can be done either by cropping out the region\cite{contrast}\cite{cf-gs}\cite{css} or by replacing it with GAN-based in-painting\cite{inv-cov}\cite{mutant}. \cite{mutant} also experiments with inverting the color of an object in the image. Similar to images, questions can be modified by masking one or more words\cite{contrast}\cite{css}\cite{mutant}. In addition to word masking, \cite{mutant} also replaces words with their negations and adversarial substitutes.

\section{Conclusion}
Notice that we did not talk at length about the details of any particular dataset or model. Our goal here was to give a bird's-eye view of the entire VQA field. So we chose not to engage in lengthy discussions of any model or dataset. In preparing this survey, we found that many outdated models still provided valuable insights or a novel way of approaching the problem. We stress that accuracy is not everything as biases in widely used datasets, VQA-v1 and v2 for example, can present a skewed depiction of a model's true ability. It would not be a stretch to say that the VQA field has exploded in a few short years. It is easy to feel overwhelmed and lost as a newcomer. We hope that our work will provide anyone new to this field with a map and a starting point.  

\bibliographystyle{plain}
\bibliography{bibliography.bib}
\EOD
\end{document}